\documentclass[sigconf]{acmart}  
\AtBeginDocument{%
  }

\setcopyright{acmlicensed}
\copyrightyear{2018}
\acmYear{2018}
\acmDOI{XXXXXXX.XXXXXXX}
\acmConference[Conference acronym 'XX]{Make sure to enter the correct
  conference title from your rights confirmation email}{June 03--05,
  2018}{Woodstock, NY}
\acmISBN{978-1-4503-XXXX-X/2018/06}

\usepackage{tcolorbox}
\usepackage{multirow}
\usepackage{multicol}
\usepackage{booktabs}
\usepackage{subfigure}
\usepackage{subcaption}
\usepackage{hhline}
\usepackage{color}
\usepackage{wasysym}
\usepackage{pifont}
\usepackage[ruled, vlined]{algorithm2e}
\usepackage{algpseudocode}
\usepackage{colortbl}
\usepackage{enumitem}
\usepackage{adjustbox}
\usepackage{graphicx}
\usepackage{soul} 

\newcommand{\OurMODEL}{\textsc{GraphPRM}}
\newcommand{\OurDataset}{\textsc{GraphSilo}}

\newcommand{\vpara}[1]{\vspace{0.05in}\noindent\textbf{#1 }}
\newcommand{\best}{\cellcolor[HTML]{EFEBF7}} 

\definecolor{bluex}{rgb}{0.27, 0.42, 0.81}
\definecolor{purplex}{HTML}{9564bf}
\definecolor{red3}{HTML}{C52A20}
\definecolor{red2}{HTML}{B36A6F}
\definecolor{red1}{HTML}{FFb5b5}
\definecolor{purple}{HTML}{B36A6F}
\definecolor{darkyellow}{HTML}{D5BA82}
\definecolor{blue1}{HTML}{508AB2}
\definecolor{blue2}{HTML}{C4E4E3}
\definecolor{green1}{HTML}{A1D0C7}
\definecolor{green2}{HTML}{BFF6BA}
\definecolor{green3}{HTML}{028100}
\definecolor{teal}{HTML}{508AB2}
\definecolor{purple1}{HTML}{e5e0f0}

\copyrightyear{2025}
\acmYear{2025}
\setcopyright{acmlicensed}\acmConference[KDD '25]{Proceedings of the 31st ACM
SIGKDD Conference on Knowledge Discovery and Data Mining V.2}{August 3--7,
2025}{Toronto, ON, Canada}
\acmBooktitle{Proceedings of the 31st ACM SIGKDD Conference on Knowledge
Discovery and Data Mining V.2 (KDD '25), August 3--7, 2025, Toronto, ON, Canada}
\acmDOI{10.1145/3711896.3737109}
\acmISBN{979-8-4007-1454-2/2025/08}

\begin{document}

\title{Rewarding Graph Reasoning Process makes LLMs more Generalized Reasoners}

\author{Miao Peng}
\authornote{Authors contributed equally to this research.}
\affiliation{%
  \institution{The Hong Kong University of Science and Technology (Guangzhou)}
  \country{Guangzhou, China}
}
\email{mpeng885@connect.hkust-gz.edu.cn}

\author{Nuo Chen}
\authornotemark[1]
\affiliation{%
  \institution{The Hong Kong University of Science and Technology (Guangzhou)}
  \country{Guangzhou, China}
}
\email{chennuo26@gmail.com}

\author{Zongrui Suo}
\affiliation{%
  \institution{The Hong Kong University of Science and Technology (Guangzhou)}
  \country{Guangzhou, China}
}
\email{zsuo785@connect.hkust-gz.edu.cn}

\author{Jia Li}
\authornote{Corresponding author.}
\affiliation{%
  \institution{The Hong Kong University of Science and Technology (Guangzhou)}
  \city{Guangzhou}
  \country{China}}
\email{jialee@ust.hk}


\begin{abstract}
Despite significant advancements in Large Language Models (LLMs), developing advanced reasoning capabilities in LLMs remains a key challenge. Process Reward Models (PRMs) have demonstrated exceptional promise in enhancing reasoning by providing step-wise feedback, particularly in the context of mathematical reasoning. However, their application to broader reasoning domains remains understudied, largely due to the high costs associated with manually creating step-level supervision. In this work, we explore the potential of PRMs in graph reasoning problems - a domain that demands sophisticated multi-step reasoning and offers opportunities for automated step-level data generation using established graph algorithms. We introduce \textbf{\OurDataset{}}, the \textbf{largest} dataset for graph reasoning problems with fine-grained \textbf{step-wise labels}, built using automated Task-oriented Trajectories and Monte Carlo Tree Search (MCTS) to generate detailed reasoning steps with step-wise labels. Building upon this dataset, we train \textbf{\OurMODEL{}}, the first PRM designed for graph reasoning problems, and evaluate its effectiveness in two key settings: \textit{inference-time scaling} and \textit{reinforcement learning} via Direct Preference Optimization (DPO). Experimental results show that \OurMODEL{} significantly improves LLM performance across 13 graph reasoning tasks, delivering a 9\% gain for Qwen2.5-7B and demonstrating transferability to new graph reasoning datasets and new reasoning domains like mathematical problem-solving. Notably, \OurMODEL{} enhances LLM performance on GSM8K and MATH500, underscoring the cross-domain applicability of graph-based reasoning rewards. Our findings highlight the potential of PRMs in advancing reasoning across diverse domains, paving the way for more versatile and effective LLMs. Codes and data are available at \url{https://github.com/GKNL/GraphPRM}.
\end{abstract}

\ccsdesc[500]{Computing methodologies~Knowledge representation and reasoning}
\ccsdesc[500]{Computing methodologies~Natural language processing}

\keywords{Large Language Models, Graph Reasoning, Process Reward Model}


\maketitle

\newcommand\kddavailabilityurl{https://doi.org/10.5281/zenodo.15553493}

\ifdefempty{\kddavailabilityurl}{}{
\begingroup\small\noindent\raggedright\textbf{KDD Availability Link:}\\
The source code of this paper has been made publicly available at \url{\kddavailabilityurl}.
\endgroup
}

\section{Introduction}

Despite remarkable progress in scaling Large Language Models (LLMs) and their performance on established benchmarks, developing advanced reasoning capabilities—particularly in domains such as mathematical problems and code generation—remains a critical area of research~\cite{training_verifiers, self_instruct, least_to_most, step_aware_verifier, LLM_judge, ToRA, Harry_Potter, SymAgent}. Chain-of-Thought (CoT)~\cite{COT, ToT} generation plays a critical role in facilitating such reasoning by breaking down complex problems into intermediary steps, thus emulating human cognitive processes. Recent research has highlighted the success of prompting LLMs with CoT examples and finetuning them using question-CoT solution pairs~\cite{scaling_relationship, GraphWiz, TimeLLaMA, G_Refer, FtG}, with the latter method proving particularly effective. 

Currently, with the release of O1/3-like models, there has been increasing attention on the viability of using reward models to provide supervisory signals within reinforcement learning~\cite{human_feedback, training_hf, DPO, PRM800K, learning_trajectory}, as a means to enhance the reasoning capabilities of LLMs, especially in optimizing test-time computing~\cite{self_consistency, scaling_test_time_compute, self_correct}.  Two main categories of reward models have been identified: Outcome Reward Models (\textbf{ORM}s)~\cite{OVM, OVM_plan}, which deliver feedback solely at the conclusion of the problem-solving process, and Process Reward Models (\textbf{PRM}s)~\cite{step_aware_verifier, process_feedback, PRM800K, Math-Shepherd}, which provide detailed feedback at every step of reasoning. The ORM evaluates the entire sequence to assign a score, while the PRM assesses each step of the reasoning process individually. PRMs have several unique advantages. They give detailed feedback by identifying where errors occur, which is crucial for reinforcement learning and correcting mistakes automatically. Moreover, PRMs mimic human evaluation processes, understanding that an intermediate error leads to a faulty conclusion, similar to human judgment. To date, PRMs have shown exceptional performance on complex reasoning by offering fine-grained supervision.

\begin{figure}[!t]
    \centering
    \includegraphics[width=0.46\textwidth]{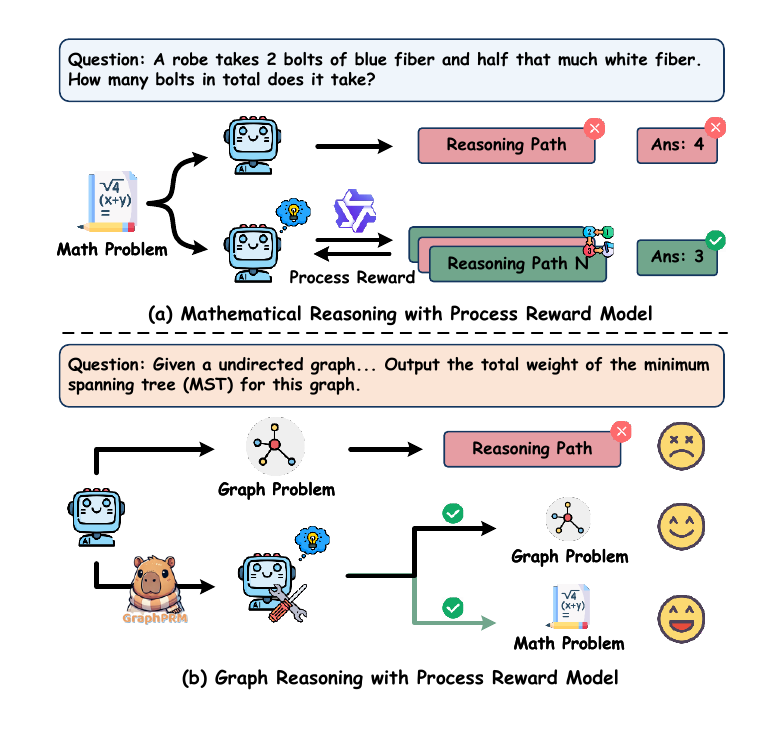}
    \caption{Example of LLM reasoning with process supervision on mathematical problems and graph reasoning problems.}
    \label{fig:intro_example}
\end{figure}

However, the exploration of PRMs has largely been confined to mathematical reasoning tasks~\cite{Math-Shepherd, fine_grained_hf}, as shown in Figure~\ref{fig:intro_example}. This narrow focus limits the understanding of how reward-based enhancements can be applied across diverse reasoning domains. Moreover, training PRMs to help mathematical problem-solving often involves constructing highly structured step-level data, which is challenging to automate due to the significant manual effort requirement~\cite{PRM800K, ProcessBench}. This scalability constraint, combined with the domain-specific nature of current PRM applications, raises important questions about their broader applicability and practical implementation across different reasoning contexts.

Graph computational problems (\textbf{GCPs}), by contrast, offer a promising avenue to overcome this limitation.
GCP reasoning~\cite{GraphWiz, GraphArena, GLBench, NLGift, ZeroG, ChapTER, PPT}
involves tackling challenges related to graph theory, a branch of mathematics concerned with the study of graph structures made up of nodes (vertices) connected by edges. GCP encompasses a wide range of complex tasks, including pathfinding, network analysis, and edge counts, which require sophisticated multi-step reasoning and the ability to navigate intricate relational structures. GCPs are intrinsically challenging for LLMs due to their dependence on spatial reasoning, the need to maintain and manipulate multiple interrelated nodes, and the requirement for precise logical deductions. Furthermore, many traditional graph algorithms already have well-established solutions. These can be leveraged to automatically generate step-level data, offering a unique opportunity to build training data for LLMs that bypasses the need for manual annotation.
This enables us to explore automated solution paths for GCPs while utilizing reward models to refine the reasoning process. 

\begin{table*}[!t]
\renewcommand{\arraystretch}{1.2}
 \centering
 \resizebox{\linewidth}{!}{
 \small
 \begin{tabular}{c|l|p{0.6\linewidth}|c|c}\toprule
    \textbf{Level} & \textbf{Task} & \textbf{Definition} & \textbf{Time Complexity} & \textbf{Answer Type}\\
    \midrule
    \multirow{5}{*}{\textbf{Node}} & \# Degree & Return the degree of node \(u\) in a given graph \(\mathcal{G}\). & $O(|\mathcal{V}|)$ & Integer \\
    & \# Clustering Coefficient & Calculate the ratio of actual edges to possible edges among node $u$'s neighbors in a graph \(\mathcal{G}\). & $O(|\mathcal{V}|^2)$ & Float \\
    & \# Neighbor & Find the neighbors of node $u$ in a given graph $\mathcal{G}$. & $O(|\mathcal{V}|)$ & Node List \\
    & \# Page Rank & Identify the node with the highest PageRank value in a given graph $\mathcal{G}$. & $O(|\mathcal{V}| + |\mathcal{E}|)$ & Integer \\
    & \# Predecessor & Return the list of predecessor nodes of $u$ in a given graph $\mathcal{G}$ & $O(|\mathcal{V}|)$ & Node List \\
    \midrule
    & \# Jaccard & Calculate the Jaccard Coefficient value $J$ between node $u$ and node $v$, where $J = \frac{|\mathcal{N}(u) \cap \mathcal{N}(v)|}{|\mathcal{N}(u) \cup \mathcal{N}(v)|}$. & $O(|\mathcal{V}|)$ & Float \\
    \textbf{Node} & \# Common Neighbor & Find the number of common neighbors of node $u$ and node $v$ in a given graph $\mathcal{G}$.& $O(|\mathcal{V}|)$& Integer \\
    \textbf{Pair} & \# Connectivity & Determine if there is a path from node $u$ to node $v$ in a given in a given graph $\mathcal{G}$. & $O(|\mathcal{V}| + |\mathcal{E}|)$ & Boolean \\
    & \# Maximum Flow & Calculate the maximum amount of flow from node $u$ to node $v$ in a given graph $\mathcal{G}$.& $O(|\mathcal{V}|^2\sqrt{|\mathcal{E}|})$ & Integer \\
    \midrule
    \multirow{4}{*}{\textbf{Graph}} & \# Breadth First Search & Return the sequence of nodes in the order of a BFS traversal starting from node $u$. & $O(|\mathcal{V}| + |\mathcal{E}|)$ & Node List \\
    & \# Cycle & Determine if there exists an cycle in a given graph $\mathcal{G}$. & $O(|\mathcal{V}| + |\mathcal{E}|)$ & Boolean \\
    & \# Diameter & Calculate the maximum distance over all pairs of nodes in a given graph $\mathcal G$. & $O(|\mathcal V| (|\mathcal{V}|+|\mathcal{E}|)\text{log}|\mathcal{V}|)$ & Integer \\
    & \# Minimum Spanning Tree & Calculate the total weight of the minimum spanning tree in a given graph $\mathcal{G}$. & $O(|\mathcal{E}|  \text{log}|\mathcal{E}|)$ & Integer \\
    \bottomrule
 \end{tabular}}
 \caption{Descriptions of graph tasks in \OurDataset{} with problem definition, time complexity of representative algorithms and answer type.  $|\mathcal V|$ and $|\mathcal E|$ indicate the number of nodes and edges in the graph $\mathcal G$.}
 \label{tab:graph_task_description}
  \vspace{-0.3cm}
\end{table*}

In light of these insights, this study explores the application of PRMs within GCP tasks to strengthen LLM. The investigation is guided by two primary research questions: First, \textit{can we train effective Graph Process Reward Models that significantly bolster LLMs' graph computational problem-solving capabilities?} Second, \textit{do these GCP-specific reward models demonstrate versatility by improving LLM performance in other complex reasoning tasks?} Addressing these questions will provide valuable insights into the scalability and cross-domain applicability of reward-based enhancements in LLMs, contributing to developing more versatile reasoning LLMs.

To this end, we first introduce the \textbf{\OurDataset{}} dataset—the foremost training dataset designed specifically for training PRMs in GCP tasks. GraphSilo represents the \textbf{largest} collection of GCPs with CoT solutions available, and it stands out as the sole process supervision dataset that incorporates detailed step-level correctness labels for GCPs. To build this dataset, we employ Monte Carlo Tree Search (MCTS)~\cite{OmegaPRM} alongside Task-oriented Trajectories, enabling the automated generation of detailed reasoning steps (More details in Section \ref{dataset}).
Building on \OurDataset{}, we then train the first process reward model tailored for GCP tasks, named \textbf{\OurMODEL{}}. We demonstrate the efficacy of \OurMODEL{} in two main settings:
1) \textbf{Inference-Time Scaling}:  \OurMODEL{} is used to rank multiple responses generated by the base LLM, selecting the best answer during inference.
2) \textbf{Reinforcement Learning}:  \OurMODEL{} helps to identify high-quality, step-wise reasoning solutions by distinguishing between correct and incorrect reasoning steps for GCPs. Positive-negative pairs are constructed based on these distinctions, with the correct reasoning steps serving as positive examples and the incorrect ones as negative examples. This pairing enables the creation of preference data for Direct Preference Optimization (DPO)~\cite{DPO}. DPO is then applied to further refine the model’s reasoning ability by optimizing the preference for high-quality solutions over suboptimal ones.

Experimentally, with the help of \OurMODEL{}, we observe significant improvements across three different backbone LLMs, including Qwen2.5~\cite{Qwen2.5}, Llama3.1~\cite{LLaMA3}, and Gemma 2~\cite{gemma2}. For example, \OurMODEL{} helps Qwen2.5-7B achieve a 9\% improvement across 13 GCP tasks during inference-time computation. Additionally, \OurMODEL{} demonstrates impressive transferability: 1) \textbf{GCP Dataset Transfer}: In zero-shot settings, it enables different LLMs to achieve a 20-30\% improvement on the GraphWiz~\cite{GraphWiz} test set across various GCP tasks, and 2) \textbf{Reasoning Domain Transfer}: Most notably, when applied to mathematical reasoning tasks, \OurMODEL{} boosts LLM performance on four math datasets, such as GSM8K and Math-500, enhancing their mathematical reasoning capabilities. These results confirm that rewarding graph reasoning processes improve LLMs’ reasoning abilities, advancing both graph computational problem-solving and other reasoning domains like mathematical problem-solving. Ultimately, our findings show that integrating graph-based reward models like \OurMODEL{} makes LLMs more versatile and effective across various reasoning tasks, bridging gaps between graph computation and general reasoning applications.

\section{Preliminaries and Related Works}
\subsection{LLMs for Graph Reasoning}
Recent research has focused on integrating LLMs with graph data due to their growing prominence. Early works like NLGraph~\cite{NLGraph} and GPT4Graph~\cite{GPT4Graph} evaluated LLMs' ability to understand graph-structured data. Subsequent studies explored fine-tuning LLMs for graph tasks, leading to models like GraphLLM~\cite{GraphLLM} and GraphGPT~\cite{GraphGPT}. Graph computational problems, requiring deeper structural understanding and multi-step reasoning, remain particularly challenging. GraphInstruct~\cite{GraphInstruct} evaluated LLMs' understanding of graphs and improved LLMs' performance on graph reasoning tasks. GraphArena~\cite{GraphArena} introduced a benchmark to test LLMs' reasoning abilities with graph computational challenges, while ProGraph~\cite{ProGraph} proposed a challenging benchmark for graph analysis using external APIs. Further advancements have focused on improving graph reasoning capabilities. GraphQA~\cite{GraphQA} showed that graph encoding significantly impacts LLM reasoning. GraphWiz~\cite{GraphWiz} improved reasoning through instruction tuning and preference alignment. Additionally, NLGift~\cite{NLGift} and GraCoRe~\cite{GraCoRe} provided evaluation suites to test generalization and reasoning with graph data. Despite significant efforts, LLMs are still struggling with solving complex graph computation problems, highlighting the need for more fine-grained, step-level process supervision in graph reasoning processes.

\subsection{Reward Models for Reasoning Problem}
Reasoning - the capacity to derive new insights, tackle challenges, and reach conclusions from given information - stands as a core capability of LLMs. Researchers have proposed various methods to improve or elicit the reasoning ability of LLMs (e.g., mathematical reasoning), which can be broadly divided into three groups: pre-training~\cite{GPT4_report, LLaMA3, Qwen2.5, gemma2}, fine-tuning~\cite{OVM_plan, WizardMath, MAmmoTH, MathOctopus, LiteCoT} and prompting~\cite{MathPrompter, self_consistency, ControlMath}. Except for directly improving and eliciting the reasoning potential of LLMs, researchers have been dedicated to boosting LLM reasoning results via an extra verifier for selecting the best answer from multiple decoded candidates. There are two primary types of verifiers: the Outcome Reward Model (ORM)~\cite{OVM} and the Process Reward Model (PRM)~\cite{PRM800K, Math-Shepherd, ProcessBench}.

\vpara{Outcome Reward Model} Given a graph computational problem $p$ and its corresponding solution $s$, ORM aims to assign a single score to $s$, which ranges in $[0,1]$ and indicates the correctness of solution $s$ towards problem $p$. Specifically, ORM is trained with a cross-entropy loss~\cite{step_aware_verifier} as follows:
\begin{equation}
\mathcal{L}_{ORM} = y_{s} \log r_{s}+\left(1-y_{s}\right) \log \left(1-r_{s}\right),
\end{equation}
where $y_{s}$ is the ground-truth of problem $p$, and $y_{s} = 1$ if solution $s$ is correct ($y_{s} = 0$ if $s$ is incorrect). $r_{s}$ indicates the sigmoid score assigned by ORM on solution $s$ from LLM generator. The performance of the ORM is critically dependent on the construction of a high-quality training dataset with outcome labels.

\vpara{Process Reward Model} Different from ORM, PRM incorporates the concept of supervising intermediate reasoning steps in addition to the final outcome, enabling more granular feedback throughout the reasoning process. Given a problem $p$ and solution $s^*$ containing multiple reasoning steps, PRM is designed to assign each step of solution $s^*$ a score ranging from 0 to 1. PRM is trained as follows:
\begin{equation}
\mathcal{L}_{P R M}=\sum_{i=1}^{N} y_{s_{i}} \log r_{s_{i}}+\left(1-y_{s_{i}}\right) \log \left(1-r_{s_{i}}\right),
\end{equation}
where $s_i$ is the $i$-th step of $s^*$, and $y_i$ is the ground-truth label of step $s_i$. $r_{s_i}$ indicates the sigmoid score of $s_i$ assigned by PRM. Likewise, the training of PRM can be regarded as a binary classification task. which means the value of $y_i$ can be 0 or 1. To obtain high-quality process labels of PRM training dataset, existing methods include 1) human annotation~\cite{PRM800K} and 2) automate annotation via Monte Carlo estimation~\cite{Math-Shepherd, OmegaPRM}.

\begin{figure*}[!t]
    \centering
    \includegraphics[width=\textwidth]{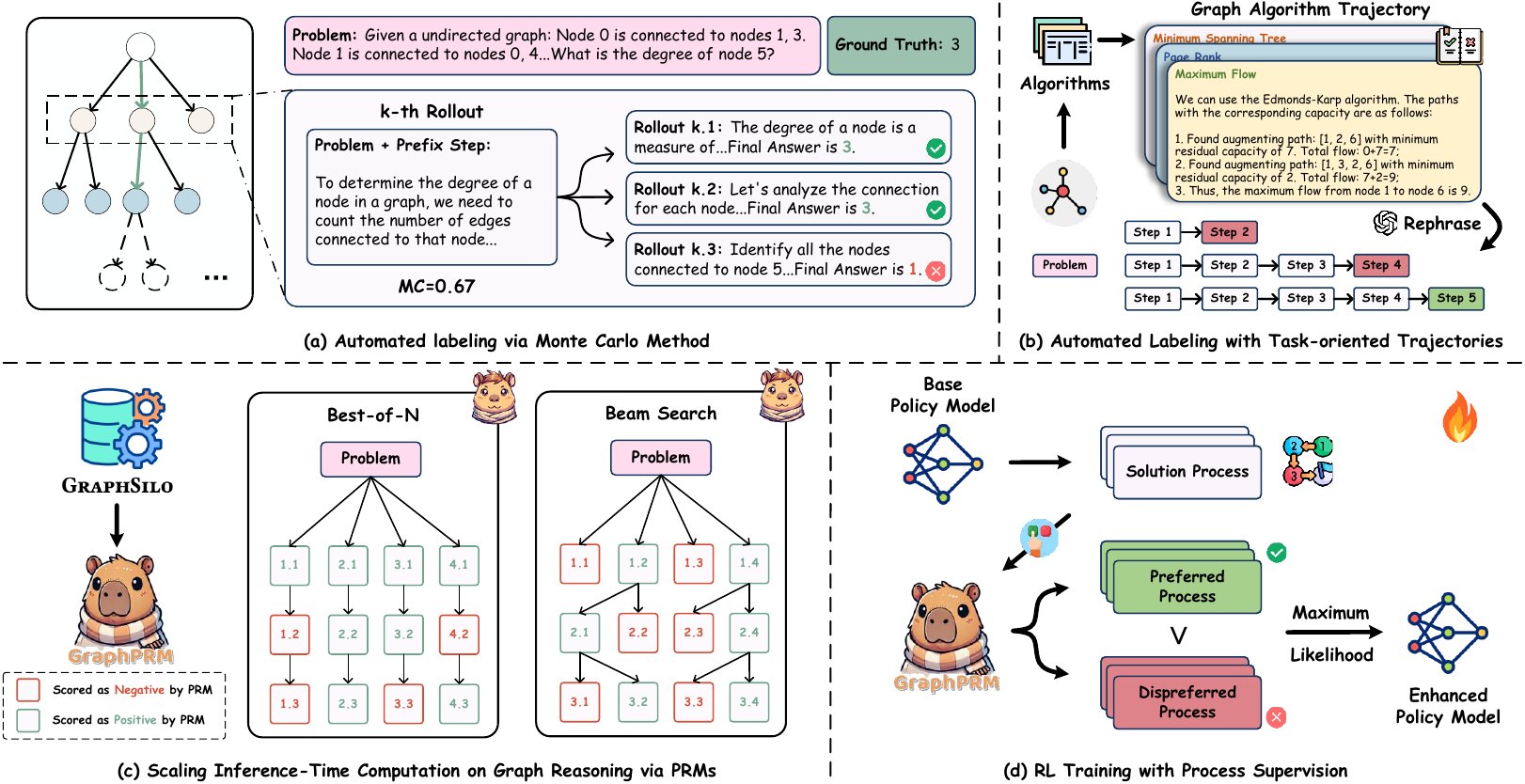}
    \caption{Overall illustration of process annotation pipeline of \OurDataset{} (a, b) and the strategies of utilizing \OurMODEL{} to strengthen LLMs on graph reasoning, including scaling inference-time computation (c) and reinforcement learning (d).}
    \label{fig:data_construction_pipeline}
\end{figure*}

\subsection{Task Formulation}

To evaluate the effectiveness of our proposed \textbf{\OurMODEL{}}, two aspects are taken into consideration: (1) \textit{Inference-Time scaling}: we utilize \OurMODEL{} to rank and select the correct one from multiple outputs from  LLMs during inference  (2) \textit{Reinforcement Learning}: \OurMODEL{} is utilized to further improve LLM in graph reasoning tasks via reinforcement learning with process supervision.

\vpara{Scaling Inference-Time Performance} 
In general, leveraging additional computation at inference time allows LLMs to surpass their training performance, unlocking new possibilities in reasoning tasks~\cite{PRM800K, scaling_test_time_compute}. By integrating a PRM, the performance of an LLM on a specific prompt can be enhanced during inference. Through multiple samples generated by the base LLM, PRMs evaluate and score individual steps in each solution, employing step-wise and inter-answer aggregation to identify the optimal answer. This approach effectively scales LLM inference performance via PRM-based search. We adopt the Best-of-N and Beam Search paradigms~\cite{scaling_test_time_compute} to demonstrate how PRMs improve LLM performance.

\vpara{Reinforcement Learning with Process Supervision} An enhanced reward model is instrumental in training higher-performing LLMs. We further use \OurMODEL{} to boost LLMs in graph reasoning via Direct Preference Optimization (DPO) with process supervision.
We utilize \OurMODEL{} to select high-quality verified step-wise reasoning solutions with positive-negative pairs for GCPs. Then, the LLM is further optimized on the preferred reasoning solutions to further sharpen its graph reasoning abilities.

\section{Methodology}

In this section, we present the construction pipeline of \OurDataset{}, including graph task selection, graph generation, and process annotation. We then describe the supervised training of \OurMODEL{}. Using the trained PRM model, we further introduce approaches of employing \OurMODEL{} to enhance LLMs by scaling inference-time computation and enabling reinforcement learning training with process supervision. An overall illustration is detailed in Figure~\ref{fig:data_construction_pipeline}.

\subsection{\OurDataset{} Construction}
\label{dataset}
To enhance LLM reasoning in solving GCPs, we create a new dataset, \OurDataset{}, for training a PRM. This section outlines the procedures including graph task selection, graph generation, process annotation, and \OurDataset{} statistics.

\subsubsection{\textbf{Graph Task Selection}}
We comprehensively consider a broad range of thirteen graph reasoning tasks in \OurDataset{}, including node level, node pair level and graph level graph computational problems. Specifically, we include five node level tasks: \textit{Degree}, \textit{Clustering Coefficient}, \textit{Neighbor}, \textit{Page Rank} and \textit{Predecessor}; four node pair level tasks: \textit{Jaccard}, \textit{Common Neighbor}, \textit{Connectivity} and \textit{Maximum Flow}; and four graph level tasks: \textit{Breadth First Search}, \textit{Cycle}, \textit{Diameter} and \textit{Minimum Spanning Tree}. The tasks are summarized in Table~\ref{tab:graph_task_description}. These thirteen graph tasks encompass extensive computational and mathematical reasoning, requiring LLMs to comprehend graph structures and perform reasoning. They facilitate a deeper theoretical understanding of graph algorithms while also addressing a wide range of practical applications.

\subsubsection{\textbf{Graph Generation}}\label{sec:graph_generation}
\OurDataset{} features diverse graphs of varying structures and sizes to evaluate LLM capabilities. Following~\cite{GraphInstruct}, we include three types of graph structure distributions: 1) random networks, 2) small-world networks~\cite{small_world_networks}, and 3) BA scale-free networks~\cite{BA_graph}. For each type, we consider both directed and undirected graphs with random generation. In terms of graph size, we include four distinct levels: Tiny (5-7 nodes), Small (8-15 nodes), Medium (16-25 nodes), and Large (26-35 nodes). These sizes represent varying levels of difficulty for LLMs in understanding and reasoning about graphs. Detailed descriptions of graph structure are presented at Appendix~\ref{app:graph_structure_details}.

\begin{figure}[!t]
    \centering
    \includegraphics[width=0.48\textwidth]{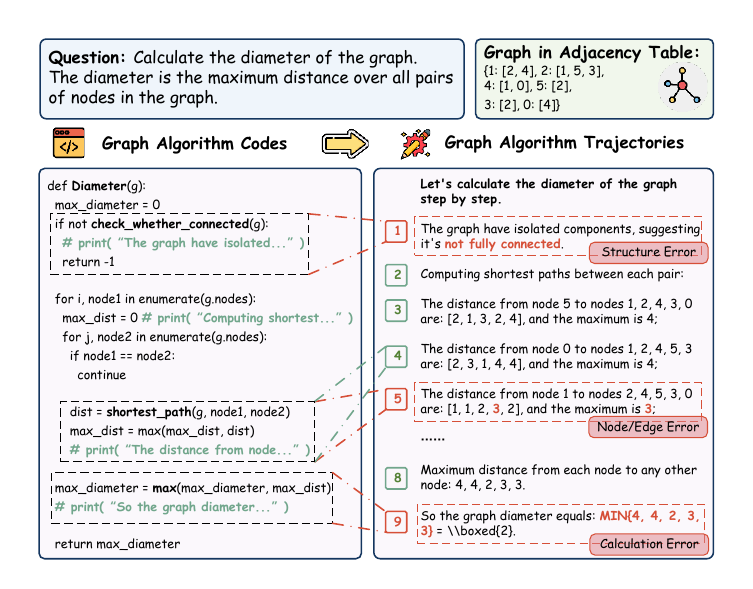}
    \caption{Illustration of generating task-oriented trajectories with step-wise labels based on graph algorithm codes.}
    \label{fig:task_oriented_trajectory}
    \vspace{-0.4cm}
\end{figure}

\subsubsection{\textbf{Process Annotation}}\label{sec:process_annotation}
Training an effective PRM for GCP tasks relies on step-level reasoning labels. Unlike prior work~\cite{PRM800K}, which depends on human annotations, GCP offers two inherent advantages: 1) their outcomes are explicit and well-defined, and 2) they involve analyzing and manipulating graph structures, in which established graph algorithms are efficiently used. Based on these, we propose our process annotation approach as follows:

\vpara{Automated labeling with Task-oriented Trajectories} Since graph algorithms follow standardized pipelines, the process of solving graph problems can naturally facilitate the generation of task-specific trajectories, which in turn produce fine-grained step-wise labels. Specifically, we select representative algorithms and corresponding codes for each GCP. Then, we translate the whole execution process of the code to trajectories in natural language, which can be further divided into multiple sub-steps, serving as positive labels.
To construct negative sub-steps, these positive sub-steps can then be perturbed using three strategies: 1) \textbf{structure-based}: introducing errors related to the graph structure (e.g., incorrect assumptions about whether the graph is directed or connected); 2) \textbf{node/edge-based}: introducing errors by randomly adding or removing nodes or edges; and 3) \textbf{calculation-based}: introducing errors through incorrect formulas or miscalculated results. 

We present a more detailed example of generating reasoning process and step-level labels in \texttt{diameter} task in Figure~\ref{fig:task_oriented_trajectory}. In this example, the trace breaks down the steps: 1) Graph Structure Check;
2) Shortest Path Calculation;
3) Distance Calculation;
4) Error Identification;
5) Diameter Calculation. We can obtain the correct steps by printing the execution process in the code (green color). Besides, we can collect negative steps by randomly introducing structure errors in graph structure check part, node/edge errors in the shortest path calculation part, and calculation errors in the diameter calculation part (red color).
With obtained reasoning processes and step-wise labels, we employ Qwen2.5-72B-Instruct to rephrase them, mimicking LLM distributions while avoiding uniform templates.

\begin{table}[t]
\renewcommand{\arraystretch}{1.4}
\centering
\small
\setlength{\tabcolsep}{1.4mm}{
\resizebox{\linewidth}{!}{
\begin{tabular}{l|ccc|cc}
\toprule
\textbf{Dataset} & \textbf{Training Set} & \textbf{CoT Process} & \textbf{Step-wise Label} & \textbf{Task} & \textbf{Size} \\ 
\midrule
NLGraph~\cite{NLGraph} & \textcolor{red}{\ding{55}} & \textcolor{red}{\ding{55}} & \textcolor{red}{\ding{55}} & 8 & 29,370 \\
GraphQA~\cite{GraphQA} & \textcolor{red}{\ding{55}} & \textcolor{red}{\ding{55}} & \textcolor{red}{\ding{55}} & 12 & 2,300 \\
GraphInstruct~\cite{GraphInstruct}  & \textcolor{green}{\ding{51}} & \textcolor{green}{\ding{51}} & \textcolor{red}{\ding{55}} & 21 & 18,900 \\
GraphWiz~\cite{GraphWiz}  & \textcolor{green}{\ding{51}} & \textcolor{green}{\ding{51}} & \textcolor{red}{\ding{55}} & 9 & 72,785 \\
NLGift~\cite{NLGift}  & \textcolor{green}{\ding{51}} & \textcolor{red}{\ding{55}} & \textcolor{red}{\ding{55}} & 4 & 37,000 \\
ProGraph~\cite{ProGraph}  & \textcolor{green}{\ding{51}} & \textcolor{green}{\ding{51}} & \textcolor{red}{\ding{55}} & 3 & 29,260 \\
GraphArena~\cite{GraphArena}  & \textcolor{red}{\ding{55}} & \textcolor{red}{\ding{55}} & \textcolor{red}{\ding{55}} & 10 & 10,000 \\
\toprule
\rowcolor{green!10} \textbf{\OurDataset{} (Ours)}  & \textcolor{green}{\ding{51}} & \textcolor{green}{\ding{51}} & \textcolor{green}{\ding{51}} & 13 & \textbf{118,189} \\
\bottomrule
\end{tabular}}}
\caption{Comparison of \OurDataset{} with related datasets.}
\label{tab:benchmark_comparison}
\vspace{-0.8cm}
\end{table}

\vpara{Automated labeling via Monte Carlo Method} To enhance the limited variety of task-oriented trajectories from graph algorithms, we use the Monte Carlo method~\cite{Math-Shepherd, OmegaPRM} to automatically generate and label graph reasoning steps. Following~\cite{OmegaPRM}, we construct a search tree for each problem-answer pair, where each node contains the problem $p$, solution prefix $s$, and and all rollouts $\{(s,r_i)\}^k_{i=1}$ derived from that node (where $r_i$ denotes the $i$-th rollout). Each edge in the tree represents either a single step or a sequence of steps originating from the node. To automatically collecting graph reasoning process with labels, we adhere to the principle of selecting the most valuable rollouts during tree search. Thus, for each node, we calculate both its value function $Q(s,r)$ and exploration term $U(s)$. The value function is defined as $Q(s, r)=\alpha^{1-\mathrm{MC}(s)} \cdot \beta^{\frac{\operatorname{len}(r)}{L}}$, where $\alpha,\beta$ and $L$ are hyperparameters, $\operatorname{len}(r)$ denotes the number of tokens in a rollout, and $\mathrm{MC}(s)$ is the Monte Carlo estimation. $Q$ is supposed to indicate how likely a rollout will be chosen for each iteration. Additionally, the exploration term is computed as: $U(s)=c_{\text{puct}} \frac{\sqrt{\sum_{i} N\left(s_{i}\right)}}{1+N(s)}$, where $N(s)$ denotes the visit count and $c_{\text{puct}}$ is a constant determining the level of exploration. During the selection phase, a rollout is popped and selected according to tree statistics based on the PUCT algorithm: $(s,r)=\arg\max_{(s,r)}[Q(s,r)+U(s)]$. A binary search is subsequently employed to locate the first error in the selected rollouts, adding rollouts with $0 < MC(s)$ to the candidate pool. All positions preceding the first error are treated as new states for further exploration.

\subsubsection{\textbf{\OurDataset{} Statistics}}
This extensive dataset provides a strong foundation for training step-wise PRM and evaluating LLMs on various GCP tasks. As shown in Table~\ref{tab:benchmark_comparison}, our dataset is the largest one to date containing CoT solutions, with \textbf{118,189} samples and \textbf{394,165} step-wise labels. It is also the only dataset that includes step-level supervision, offering a valuable resource for graph reasoning research. We report detailed dataset statistics in Table~\ref{tab:Benchmark Statistics}.

\subsection{Supervised Training of \OurMODEL{}}
PRMs evaluate whether a reasoning subsequence is on the correct path by assigning a correctness score. The PRM is trained via supervised fine-tuning on an LLM using correct/incorrect labels.
PRM800K~\cite{PRM800K} classifies each step as 'good', 'neutral', or 'bad'. However, for graph reasoning tasks, the difference between binary and three-class classification is minimal, so we treat PRM training as a binary classification task. 

Specifically, we train \textbf{\OurMODEL{}} on \OurDataset{} using step-wise process annotations to evaluate the correctness of individual reasoning steps for a given GCP. Training is conducted through supervised fine-tuning on an LLM with correct/incorrect labels. Reasoning processes are divided into steps, separated by a special step token ``\textbackslash n\textbackslash n\textbackslash n\textbackslash n\textbackslash n" to indicate the end of a step, where \OurMODEL{} makes predictions. Each step is labeled `+' for correct or `-' for incorrect based on solution validity. During training, the data is formatted as a next-token prediction task, where the model learns to predict a positive or negative token following each step token. The input includes the graph computational question combined with the reasoning process, with steps separated by step tokens. To ensure \OurDataset{} evaluates the effectiveness of the current step within the full reasoning context, we concatenate the problem $p$ with all preceding steps ($s_1, ..., s_i$) instead of considering only the current step ($s_i$) in isolation. Labels are assigned at step token positions as either positive or negative, while other positions are ignored during loss computation. Additionally, the attention mask assigns a value of 1 to all tokens except step token positions, ensuring the LLM focuses on the input sequence without attending to step tokens. Formally, the training objective for \OurMODEL{} is defined as:
\begin{equation}
    \min_{\theta} \mathcal{L}(\theta) = - \frac{1}{N} \sum_{t=1}^T m_t \Big[ y_t \log \hat{y}_t + (1 - y_t) \log (1 - \hat{y}_t) \Big],
\end{equation}
where $\theta$ is the parameter of \OurMODEL{}, $y_i$ is the $i$-th reasoning step label and $\hat{y}_i = P(y_i \mid p, s_1, \dots, s_i)$ is the predicted probability of $i$-th step. $m_t$ indicates whether each token is the step token.

\subsection{Scaling Inference-Time Compute via PRMs}

\subsubsection{\textbf{Inference-time Searching Strategies}}
Following prior work \cite{PRM800K, scaling_test_time_compute}, we use \OurMODEL{} to evaluate individual steps in solutions from the policy model. We explore two paradigms: (1) Best-of-N selection and (2) Guided beam search, integrating PRM for guided decoding and scoring or voting across solutions.

\vpara{Best-of-N Selection} The Best-of-N search strategy involves sampling $N$ candidate solutions independently from the base LLM and selecting the most optimal solution based on a PRM's evaluation. Specifically, for a given problem $p$ in the test set, $N$ candidate solutions $s^*$ are generated by the LLM, and each step $s_i$ in solution $s^*$ is scored by \OurMODEL{}. With answer aggregation approach, we can assess the quality of each solution. The solution with the highest score is chosen as the final answer. By leveraging \OurMODEL{}, this approach increases the likelihood of selecting a correct solution, thereby enhancing the success rate of LLMs in solving GCPs.

\vpara{Guided Beam Search} Guided beam search uses PRM for step-by-step optimization. Following BFS-V~\cite{ToT}, the LLM generates $N$ candidates for the first step, which are then evaluated and scored by the PRM. The top $K (K \leq N)$ candidates with the highest scores are retained for the current step. For each of these $K$ outputs, the base LLM generates $M = N / K$ (where $M \in \mathbb{Z}$) subsequent candidates, reconstituting a total pool of $N$ outputs. This iterative process continues, with new candidates being scored, filtered, and sampled at each step. Crucially, the PRM’s scores serve as the primary mechanism for guiding and refining the search trajectory at each step.

\subsubsection{\textbf{Answer Aggregation}}
During inference, \OurMODEL{} scores each step of a solution from the base LLM. To identify the final answer, a two-step aggregation function is required to aggregate per-step scores to determine the most likely correct solution. First, we combine the per-step scores to compute a final score for the entire single solution. Following~\cite{PRM800K, scaling_test_time_compute}, we adopt two approaches to aggregate step-wise scores for a single solution $s$:
1) \textbf{PRM-Last}: We take the score of the last step $s_e$ as the full-solution score, denoted as $v_{s^*} = r_{s_e}$.
2) \textbf{PRM-Min}: Choose the minimum score among all steps to be the overall score, denoting as $v_{s^*} = \min \left( r_{s_1}, r_{s_2}, \dots, r_{s_e} \right)$, which is demonstrated as good as \textbf{PRM-Last} strategy~\cite{scaling_test_time_compute}. After that, we aggregate the final scores across all solutions. Through a \textbf{Weighted Majority Vote}, combining self-consistency and PRM, solutions are grouped by their final answers, and an aggregate score is computed for each group to select the best overall answer.
Formally, the final answer based on $N$ candidate solutions is:
\begin{equation}
    z_{s^*}=\underset{z}{\arg \max } \sum_{i=1}^{N} \mathbb{I}\left(z_{i}=z\right) \cdot v_{s^*},
\end{equation}
where $z_i$ represents the answer of $i$-th generated solution, and $v_{s^*}$ denotes the aggregated score of solution $s^*$, which can be implemented as either PRM-Last or PRM-Min.
Notably, we denote the above combined two multi-answer weighting methods as \textbf{PRM-Last-Vote} and \textbf{PRM-Min-Vote}.

\begin{table*}[!t]\small
\renewcommand{\arraystretch}{1.1}
 \centering
 \setlength{\tabcolsep}{0.8mm}
 \resizebox{\linewidth}{!}{
 \begin{tabular}{l|cccccccccc|c|ccc|c}\toprule
    \multirow{2}{*}{\textbf{Verification Strategy}} & \multicolumn{10}{c|}{\textbf{In-domain Tasks}} & \multirow{2}{*}{\textbf{Average}} & \multicolumn{3}{c|}{\textbf{Out-of-domain Tasks}} & \multirow{2}{*}{\textbf{Average}}  \\
    \cmidrule(lr){2-11}\cmidrule(lr){13-15}
    & \multicolumn{1}{c|}{Degree} & \multicolumn{1}{c|}{Clustering} & \multicolumn{1}{c|}{Jaccard} & \multicolumn{1}{c|}{Common} & \multicolumn{1}{c|}{Connectivity} & \multicolumn{1}{c|}{Diameter} & \multicolumn{1}{c|}{Page Rank} & \multicolumn{1}{c|}{MST} & \multicolumn{1}{c|}{Flow} & \multicolumn{1}{c|}{Predecessor} & & \multicolumn{1}{c|}{Neighbor} & \multicolumn{1}{c|}{BFS} & Cycle \\
    \midrule \specialrule{0em}{1.5pt}{1.5pt}
    \multicolumn{1}{c}{\textbf{\textit{Qwen2.5-1.5B-Instruct}}} \\
    Self-Consistency     & 84.33 & 9.00 & 32.33 & 44.67 & 31.00 & 13.33 & 8.33 & 0.33 & \best{6.33} & \best{3.67} & 23.33 & 33.33 & 8.33 & 50.33 & 30.67 \\
    \OurMODEL{} (Best-of-N)     & 86.67 & 9.00 & 37.00 & \best{45.67} & 32.67 & 12.33 & 10.00 & 0.33 & 5.67 & 3.33 & 24.27 & 39.67 & \best{9.67} & 50.33 & 33.22 \\
    \OurMODEL{} (Beam Search)     & \best{92.67} & \best{13.00} & \best{53.00} & 45.00 & \best{67.33} & \best{15.00} & \best{10.67} & \best{1.00} & 3.67 & 2.33 & \best{\textbf{30.37}} & \best{52.33} & 7.00 & \best{51.00} & \best{\textbf{36.78}} \\
    \midrule
    \multicolumn{1}{c}{\textbf{\textit{Qwen2.5-7B-Instruct}}} \\
    Self-Consistency     & 90.00 & 25.33 & 58.00 & 49.00 & 43.33 & \best{48.00} & 12.33 & \best{0.33} & 13.33 & 30.67 & 37.03 & 69.67 & 48.00 & 52.33 & 56.67 \\
    \OurMODEL{} (Best-of-N)     & 90.67 & \best{26.33} & 60.00 & \best{49.33} & 49.67 & 46.67 & \best{13.67} & \best{0.33} & \best{14.00} & 32.33 & 38.30 & 73.33 & \best{49.00} & 53.00 & \best{\textbf{58.44}} \\
    \OurMODEL{} (Beam Search)     & \best{94.33} & 22.33 & \best{74.33} & 48.00 & \best{79.33} & 43.33 & 11.33 & 0.00 & 7.33 & \best{35.33} & \best{\textbf{41.57}} & \best{86.00} & 33.67 & \best{53.33} & 57.67 \\
    \midrule
    \multicolumn{1}{c}{\textbf{\textit{LLaMA3.1-8B-Instruct}}} \\
    Self-Consistency     & 84.00 & 10.00 & 44.00 & 46.33 & 23.00 & 17.33 & 15.33 & 0.00 & 6.67 & 12.67 & 25.93 & 61.00 & 26.00 & 49.67 & 45.56 \\
    \OurMODEL{} (Best-of-N)     & 87.00 & 12.67 & 43.33 & 46.33 & 24.00 & 18.00 & \best{16.33} & 0.33 & 7.00 & 12.33 & 26.73 & 65.00 & \best{26.67} & \best{50.00} & \best{\textbf{47.22}} \\
    \OurMODEL{} (Beam Search)     & \best{94.33} & \best{14.33} & \best{62.33} & \best{47.33} & \best{37.00} & \best{26.33} & 9.33 & \best{1.00} & \best{7.33} & \best{14.00} & \best{\textbf{31.33}} & \best{78.00} & 18.33 & 38.33 & 44.89 \\
    \midrule
    \multicolumn{1}{c}{\textbf{\textit{Gemma2-9B-Instruct}}} \\
    Self-Consistency     & 95.67 & 10.00 & 41.00 & 33.33 & 55.00 & 31.67 & \best{13.00} & 0.00 & 7.67 & 30.33 & 31.77 & 93.67 & 31.67 & 54.33 & 59.89 \\
    \OurMODEL{} (Best-of-N)     & 97.00 & 10.67 & 41.67 & 34.00 & 59.33 & \best{33.67} & 12.67 & 0.00 & 7.67 & \best{30.67} & 32.73 & \best{94.33} & \best{35.67} & \best{59.67} & \best{\textbf{63.22}} \\
    \OurMODEL{} (Beam Search)     & \best{97.67} & \best{12.67} & \best{43.00} & \best{39.00} & \best{77.67} & 25.33 & 9.67 & 0.00 & \best{8.67} & 28.00 & \best{\textbf{34.17}} & 90.00 & 35.33 & 56.67 & 60.67 \\
    \bottomrule
 \end{tabular}}
 \caption{Performances (\%) of different LLMs on \OurDataset{} in both in-domain and out-of-domain tasks with different verification strategies. The reward model is trained with Qwen2.5-Math-7B on \OurDataset{}. For both Best-of-N and Beam Search strategies, the generation budget is set to 8 and the answer aggregation strategy is PRM-Last. The best results are in \sethlcolor{purple1}\hl{Purple}.}
 \label{tab:main_results_graphsilo}
  \vspace{-0.4cm}
\end{table*}

\subsection{RL Training with Process Supervision}
Upon achieving \OurMODEL{}, we employ reinforcement learning to further train LLMs. Considering the instability of PPO~\cite{PPO} training, we choose Direct Preference Optimization (DPO)~\cite{DPO} instead. Under this circumstance, \OurMODEL{} guides the selection of high-quality preference pairs, ensuring correctness in preferred solutions and falsehood in dispreferred ones. Thus, DPO refines the original policy model by aligning its parameter $\pi(\theta)$ more closely with preferred outcomes using a training corpus similar to the distribution from the policy model. To be specific, DPO employs input pairs labeled $(\mathcal{S}_w,\mathcal{S}_l)$, where $\mathcal{S}_w$ and $\mathcal{S}_l$ are the set of preferred and less preferred reasoning processes. To obtain these training data, we first employ Supervised Fine-Tuning (SFT) to train a base LLM with language modeling loss on all tasks from \OurDataset{}. The model is trained to process graph $\mathcal{G}$ and problem $\mathcal{P}$ as input and generate reasoning process $S$ as output, obtaining model $\mathcal{M}(\theta)_{sft}$ as follows:
\begin{equation}
\begin{array}{l}
\mathcal{L}_{\mathrm{LM}}=-\sum_{i=1}^{N} \sum_{j=1}^{M} \log \mathrm{P}\left(\mathcal{S}_{i, j} \mid \mathcal{G}_{i}, P_{i} ; \theta\right),
\end{array}
\end{equation}
where $S$ is the set of reasoning processes towards graph problems, $N$ denotes the number of problems and $M$ represents the number of reasoning processes for each graph problem.
Then we generate a new set of 20,000 graph problems, and perform two strategies on $\mathcal{M}(\theta)_{sft}$ to obtain the preference dataset: 1) By performing Beam Search with 8 width on $\mathcal{M}(\theta)_{sft}$ with our trained \OurMODEL{}, we take the best output of each problem as its preferred reasoning process $\mathcal{S}_w$. 2) We apply Best-of-N on $\mathcal{M}(\theta)_{sft}$ with \OurMODEL{}, generating 8 samples per problem. The sample with the wrong answer and lowest step score is chosen as the dispreferred reasoning process $\mathcal{S}_l$. Through this we can obtain fine-grained preference dataset $\mathcal{D}$ with process supervision, resulting in 4,716 $(\mathcal{S}_w,\mathcal{S}_l)$ pairs. The formal training objective for DPO is defined as follows:
\begin{equation}
\begin{array}{l}
\mathcal{L}_{D P O}\left(\mathcal{M}(\theta) ; \mathcal{M}(\theta)_{\mathrm{sft}}\right)=-\mathbb{E}_{\left(x, \mathcal{S}_{w}, \mathcal{S}_{l}\right) \sim D} \\
{\left[\log \sigma\left(\beta \log \frac{\mathcal{M}(\theta)\left(\mathcal{S}_{w} \mid x\right)}{\mathcal{M}(\theta)_{\mathrm{sft}}\left(\mathcal{S}_{w} \mid x\right)}-\beta \log \frac{\mathcal{M}(\theta)\left(\mathcal{S}_{l} \mid x\right)}{\mathcal{M}(\theta)_{\mathrm{sft}}\left(\mathcal{S}_{l} \mid x\right)}\right)\right],}
\end{array}
\end{equation}
where $x$ is the concatenation of problem and reasoning processes and $\beta$ is the preference hyper-parameter.

\section{Experiments}

In this section, we evaluate the proposed \OurMODEL{} by addressing the following research questions:
\begin{itemize}
\item  \textbf{RQ1:} How does \OurMODEL{} scales inference-time performance over LLMs on \OurDataset{}, particular when transferring to unseen and out-of-dataset graph tasks?
\item  \textbf{RQ2:} How transferable is \OurMODEL{} to mathematical reasoning problems?
\item  \textbf{RQ3:} How does \OurMODEL{} benefit LLMs reasoning abilities via Preference Alignment with process supervision?
\item  \textbf{RQ4:} How does graph size affect the performance of \OurMODEL{} on LLMs during inference-time?
\item  \textbf{RQ5:} How does process label quality affect \OurMODEL{}?
\item  \textbf{RQ6:} What is the influence of parameter sizes on the interplay between policy model and \OurMODEL{}?
\end{itemize}

More results and analysis about \OurMODEL{} on different models and graph densities are presented at \textbf{Appendix~\ref{app:additional_experimental_results}}.

\subsection{Experimental Settings}
\vpara{Datasets} We conduct experiments on both \textbf{graph reasoning} and \textbf{math reasoning} tasks. For graph reasoning, we use our \OurDataset{} alongside the widely used GraphWiz~\cite{GraphWiz} as the test set. Tasks in \OurDataset{} are divided into In-domain and Out-of-domain groups based on their presence in \OurMODEL{}'s training data. We designate BFS, Cycle, and Neighbors as Out-of-domain tasks because: 1) These three tasks are relatively challenging, requiring LLMs to possess a certain level of reasoning abilities to complete; 2) They cover both node-level and graph-level aspects, and outputs involve diverse formats.
For math reasoning, we evaluate widely-used datasets, including GSM8K~\cite{training_verifiers}, GSM8K-Hard~\cite{PAL}, MATH500~\cite{MATH}, and SVAMP~\cite{SVAMP}. We evaluate the LLM's output using Exact-Match accuracy~\cite{scaling_test_time_compute}. For float outputs, a relative error of less than 1e-4 is required. Evaluations use a zero-shot prompt setting, with results reported as mean accuracy over three sampled groups. In the reinforcement learning setting, the entire test set is used to evaluate model performance.

\vpara{Setting on Inference-time Scaling} Experiments are based on a series of large language models including Qwen2.5-1.5B/7B, Qwen2.5-Math-1.5B/7B~\cite{Qwen2.5}, LLaMA3.1-8B~\cite{LLaMA3} and Gemma2-9B\cite{gemma2}. We train our \OurMODEL{} on Qwen2.5-Math-7B-Instruct with 118k step-level data from \OurDataset{}. \OurMODEL{} is trained for three epochs with a learning rate of 1e-4. During inference-time computation, we set the decode number $N$ to 8 and temperature parameter to 0.7 for both searching strategies in most experiments. Among different aggregation strategies, we select PRM-Last-Vote as a representative. The LLM inference server is implemented using FastChat~\cite{LLM_judge, wang2024openr}.

\vpara{Setting on Reinforcement Learning} In the reinforcement learning experiments, we use the Qwen2.5-7B-Instruct model as the base model for SFT and DPO training, with the trained \OurMODEL{} serving as the PRM to select and construct the preference dataset for DPO training. For SFT training, we construct a new set of graph tasks in \OurDataset{} containing 20,000 problem-process pairs. In DPO training, the learning rate is 5e-6, and the preference hyperparameter $\beta$ is set to 0.1.

\subsection{Performance of \OurMODEL{} (RQ1)}\label{sec:performance_of_GraphPRM}
To evaluate \OurMODEL{}, we conduct inference-time scaling experiments with Best-of-N and Beam Search strategies on \OurDataset{}, and performance is reported in Table~\ref{tab:main_results_graphsilo}. Notably, compared to Self Consistency, LLMs with \OurMODEL{} achieve remarkable improvements during inference-time computation with the same generation budget, especially under the Beam Search strategy via \OurMODEL{}. This consistency is maintained across a range of tasks at different levels. Besides, \OurMODEL{} achieves a general better performance transferring to those out-of-domain tasks (Neighbor, BFS, Cycle) that are not included in the training dataset of \OurMODEL{}.

\vpara{Comparison against ORMs/Math-PRMs} As \OurMODEL{} is the first PRM specific for GCP tasks, to further demonstrate the effectiveness of \OurMODEL{} compared to ORM, we train an additional \textbf{G-ORM} on GraphSilo without step-wise labels for GCP tasks (only use outcome reward for supervision). As reported in Table~\ref{tab:main_results_ORM}, we can observe that though G-ORM outperforms SC, it lags behind \OurMODEL{} on both in-domain and out-domain graph reasoning tasks, which further prove the effectiveness of \OurMODEL{}. Beyond this, we also investigate the performance of representative math-specific PRMs on \OurDataset{}, and Table~\ref{tab:main_results_ORM} shows that math-specific PRMs cannot easily transfer to graph reasoning tasks. We conclude this phenomenon for the following reasons: 1) Graph reasoning tasks involve diverse reasoning patterns (e.g., topological, logical, and computational reasoning) that inherently require an understanding of graph structures, which is less pronounced in mathematical problem-solving. 2) The difficulty of graph problems often scales exponentially with the size and complexity of the graph.

\begin{table}[!t]\small
\renewcommand{\arraystretch}{1.1}
 \centering
 \setlength{\tabcolsep}{0.9mm}
 \resizebox{0.9\linewidth}{!}{
 \begin{tabular}{l|c|c|c}\toprule
    \textbf{Methods} & \textbf{In-domain} & \textbf{Out-domain} & \textbf{Average}  \\
    \midrule \specialrule{0em}{1.5pt}{1.5pt}
    Self-Consistency     & 37.03 & 56.67 & 41.56 \\
    G-ORM     & 37.40 & 56.33 & 41.77 \\
    G-ORM + Self-Consistency     & 37.77 & 57.00 & 42.21 \\
    \midrule
    Math-Shepherd-7B~\cite{Math-Shepherd}     & 36.23 & 54.89 & 40.54 \\
    Math-PSA-7B~\cite{wang2024openr}     & 37.30 & 56.44 & 41.72 \\
    \midrule
    \OurMODEL{} (Best-of-N)     & 38.30 & \best{58.44} & 42.95 \\
    \OurMODEL{} (Beam Search)     & \best{41.57} & 57.67 & \best{45.28} \\
    \bottomrule
 \end{tabular}}
 \caption{Comparison among \OurMODEL{}, G-ORM and math-specific PRMs on \OurDataset{}. The policy model is Qwen2.5-7B-Instruct. The generation budget during verification is set to 8. The best results are in \sethlcolor{purple1}\hl{Purple}.}
 \label{tab:main_results_ORM}
  \vspace{-0.4cm}
\end{table}

\begin{table}[!t]\small
\renewcommand{\arraystretch}{1.1}
 \centering
 \setlength{\tabcolsep}{0.8mm}
 \resizebox{\linewidth}{!}{
 \begin{tabular}{l|cc|c|cc|c}\toprule
    \multirow{2}{*}{\textbf{Verification Strategy}} & \multicolumn{2}{c|}{\textbf{Easy}}  & \multicolumn{1}{c|}{\textbf{Medium}} & \multicolumn{2}{c|}{\textbf{Hard}} & \multirow{2}{*}{\textbf{Average}}  \\
    \cmidrule(){2-6}
    & \multicolumn{1}{c|}{Connect} & \multicolumn{1}{c|}{Topology} & \multicolumn{1}{c|}{shortest} & \multicolumn{1}{c|}{Hamilton} & \multicolumn{1}{c|}{Subgraph} \\
    \midrule \specialrule{0em}{1.5pt}{1.5pt}
    \multicolumn{1}{c}{\textbf{\textit{Qwen2.5-7B-Instruct}}} \\
    Self-Consistency     & 29.75 & \best{15.00} & \best{18.00} & 0.25 & 9.50 & 14.50 \\
    \OurMODEL{} (Best-of-N)     & 32.50 & 13.25 & 17.00 & 10.00 & \best{25.25} & 15.80 \\
    \OurMODEL{} (Beam Search)     & \best{34.50} & 8.25 & 14.00 & \best{29.75} & 14.00 & \best{20.10} \\
    \midrule
    \multicolumn{1}{c}{\textbf{\textit{LLaMA3.1-8B-Instruct}}} \\
    Self-Consistency     & 57.00 & 6.50 & 8.25 & 42.25 & 41.00 & 31.00 \\
    \OurMODEL{} (Best-of-N)     & 60.50 & 7.25 & 8.50 & 44.50 & 43.50 & 32.85 \\
    \OurMODEL{} (Beam Search)     & \best{82.75} & \best{12.25} & \best{10.50} & \best{51.25} & \best{57.25} & \best{42.80} \\
    \bottomrule
 \end{tabular}}
 \caption{Performances (\%) of \OurMODEL{} on GraphWiz dataset in five representative tasks. The generation budget during verification is set to 8. The best results are in \sethlcolor{purple1}\hl{Purple}.}
 \label{tab:main_results_graphwiz}
  \vspace{-0.4cm}
\end{table}

\vpara{Dataset Transfer} In addition to our \OurDataset{}, we also evaluate \OurMODEL{} on the widely used GraphWiz dataset. We select five most representative tasks (Connect, Topology, Shortest, Hamilton, and Subgraph) across three different difficulty levels and directly test \OurMODEL{} without using example prompts or additional training. As shown in Table~\ref{tab:main_results_graphwiz}, \OurMODEL{} demonstrates significant improvements in LLMs through inference-time computation, highlighting its effectiveness across datasets with diverse distributions.

\begin{figure}[!t]
    \centering
    \includegraphics[width=0.44\textwidth]{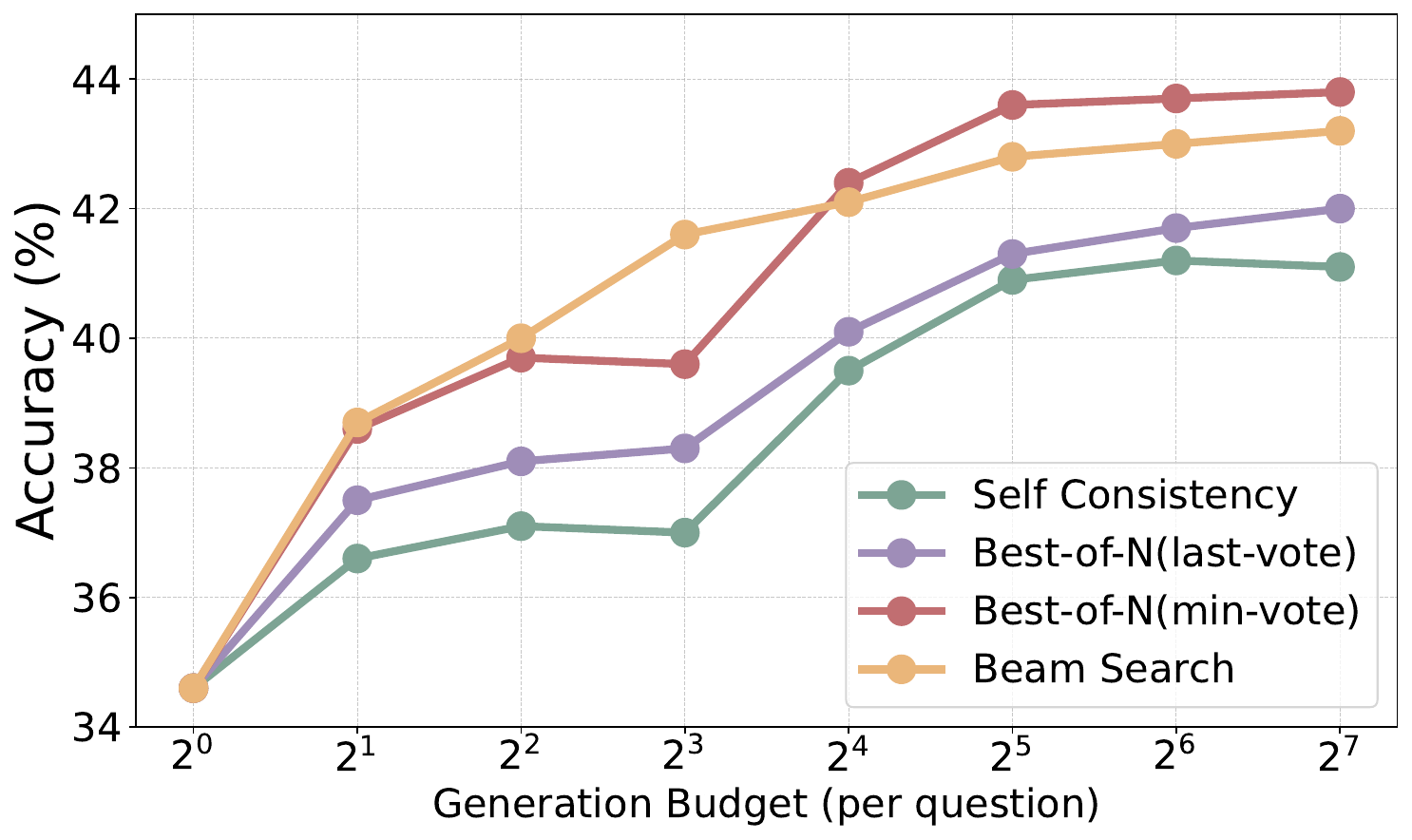}
    \caption{Performance of \OurMODEL{} with different evaluation paradigms. The generation budget ranges from 1 to 128.}
    \label{fig:silo_line_plot_all_strategy}
    \vspace{-0.4cm}
\end{figure}

\vpara{Generation Budget} We further evaluate \OurMODEL{} on LLMs with varying numbers of generation samples to analyze the impact of the generation budget (representing computational effort or token usage per question) on inference-time scaling. As shown in Figure~\ref{fig:silo_line_plot_all_strategy}, both the Best-of-N and Beam-Search methods significantly outperform Self Consistency, particularly as the generation budget increases. This confirms that \OurMODEL{} enhances LLM performance during inference-time computation, with improvements closely tied to the generation budget.

\subsection{Transferability on Math Problems (RQ2)}

\begin{figure*}[!t]
    \centering
    \includegraphics[width=\textwidth]{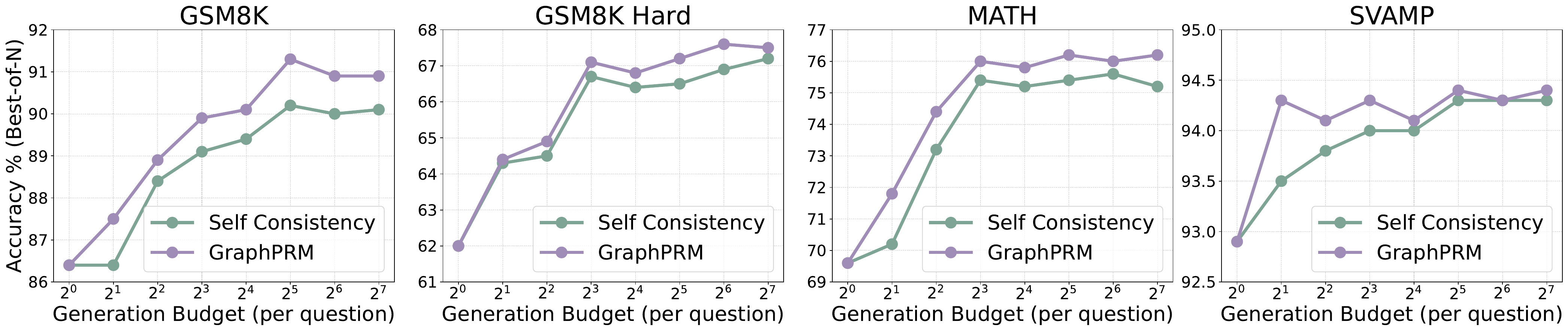}
    \caption{Best-of-N Performance of \OurMODEL{} on four math datasets with ranging generation budget (number of solution candidates). The policy model is Qwen2.5-Math-1.5B-Instruct and the answer aggregation strategy is PRM-Last.}
    \label{fig:prm_math_line_plot}
\end{figure*}

To investigate the zero-shot transferability of \OurMODEL{} on mathematical reasoning tasks, we conduct experiments on four classic math datasets. Figure~\ref{fig:prm_math_line_plot} illustrates the performance under the Best-of-N strategy during inference-time computation with varying generation budgets. It is clear that \OurMODEL{} significantly enhances the performance of the policy model on mathematical reasoning tasks and consistently outperforms the Self-Consistency method across all generation budgets. Besides, the performance gap between \OurMODEL{} and Self Consistency is more pronounced in datasets with higher complexity (e.g., GSM8K Hard and MATH) compared to simpler datasets (e.g., SVAMP).

Regarding the phenomenon that \OurMODEL{} achieves significant zero-shot transferability on mathematical reasoning tasks, we attribute it to the following reasons:
1) Graph algorithms are fundamentally grounded in mathematical concepts, such as graph theory and combinatorics, requiring models to understand complex relational modeling and optimization logic, which align closely with the demands of mathematical reasoning. 2) The abstraction inherent in graph algorithms—dealing with nodes, edges, and their relationships—enhances the model's capacity to process variables and sets in mathematical reasoning. Additionally, the recursive and divide-and-conquer strategies learned in graph algorithms (e.g., depth-first search or dynamic programming) can be directly applied to mathematical reasoning tasks, such as solving recursive relations or constructing complex proofs. 3) Graph algorithm tasks often involve dynamic inputs (e.g., variable graph structures, changing edge weights), requiring models to flexibly adjust their reasoning processes. This dynamic adaptability is equally important when dealing with open-ended and variable mathematical problems.

\vpara{Comparison against Math-PRMs on math datasets} We further compare \OurMODEL{} with two representative math-specific PRMs: Math-Shepherd-Mistral-7B~\cite{Math-Shepherd} and MATH-PSA-7B~\cite{wang2024openr}. Results reported in Table~\ref{tab:PRMs_on_Math} shown that \OurMODEL{} yield strong and competitive performance on math reasoning benchmarks. With the observation in Table~\ref{tab:main_results_ORM} that math-specific PRMs cannot easily transfer to graph reasoning tasks, it further demonstrates the uniqueness and effective transferability of \OurMODEL{}.

\begin{table}[!t]\small
\renewcommand{\arraystretch}{1.1}
 \centering
 \setlength{\tabcolsep}{0.8mm}
 \resizebox{0.9\linewidth}{!}{
 \begin{tabular}{l|c|c|c|c}\toprule
    \multirow{2}{*}{\textbf{Methods}} & \multicolumn{2}{c|}{\textbf{GSM8K}}  & \multicolumn{2}{c}{\textbf{MATH}} \\
    \cmidrule(){2-5}
    & \multicolumn{1}{c|}{Best-of-N} & \multicolumn{1}{c|}{Beam Search} & \multicolumn{1}{c|}{Best-of-N} & \multicolumn{1}{c}{Beam Search} \\
    \midrule \specialrule{0em}{1.5pt}{1.5pt}
    Self-Consistency     & 89.08 & 89.08 & 75.40 & 75.40 \\
    Math-Shepherd-7B     & 90.09 & 89.08 & 75.90 & 76.00 \\
    Math-PSA-7B     & \best{90.50} & 90.25 & \best{76.30} & 75.40 \\
    \midrule
    \OurMODEL{}     & 89.92 & \best{90.54} & 76.00 & \best{76.30} \\
    \bottomrule
 \end{tabular}}
 \caption{Comparison among \OurMODEL{} and math-specific PRMs on math datasets. The policy model is Qwen2.5-Math-1.5B-Instruct. The generation budget during verification is set to 8. The best results are in \sethlcolor{purple1}\hl{Purple}.}
 \label{tab:PRMs_on_Math}
  \vspace{-0.4cm}
\end{table}

\subsection{Performance of Post-trained LLMs with Process Supervision (RQ3)}
We report the performance of post-trained LLMs with process supervision via \OurMODEL{} in Table~\ref{tab:main_results_dpo}. "DPO-Outcome" refers to a model post-trained on a preference dataset where preferred and dispreferred reasoning processes for each problem are randomly selected based on their outcome answers (e.g., a reasoning process with a correct answer is randomly selected from all generated candidates as the preferred one). We perform DPO training on 1) vanilla Qwen2.5-7B model and 2) Qwen2.5-7B model supervised fine-tuned on all 9 training tasks from \OurDataset{}, and results indicate that DPO with process supervision via \OurMODEL{} significantly enhances the performance of two policy models, achieving improvements of 12.4\% and 10.91\%, respectively, and outperforming the "DPO-Outcome" ablation. Furthermore, integrating reinforcement learning with inference-time scaling using \OurMODEL{} proves complementary, delivering an additional 1\% performance boost.

\begin{table}[!t]\small
\renewcommand{\arraystretch}{1.0}
 \centering
 \setlength{\tabcolsep}{0.8mm}
 \resizebox{\linewidth}{!}{
 \begin{tabular}{p{0.8\linewidth}|c}\toprule
    \textbf{Models} & \textbf{\OurDataset{}} \\
    \midrule \specialrule{0em}{1.5pt}{1.5pt}
    \textbf{Qwen2.5-7B-Instruct}     & 34.60 \\
    \midrule
    + DPO - Outcome     & 35.47 \\
    + DPO - Process Supervision (Ours)     & 43.00 \\
    + DPO - Process Supervision + \OurMODEL{} (Best-of-8)     & \best{44.20} \\
    \midrule
    \midrule
    \textbf{Qwen2.5-7B-Instruct: SFT}     & 53.57 \\
    \midrule
    + DPO - Outcome     & 55.77 \\
    + DPO - Process Supervision (Ours)     & 58.47 \\
    + DPO - Process Supervision + \OurMODEL{} (Best-of-8)     & \best{59.03} \\
    \bottomrule
 \end{tabular}}
 \caption{Performances of reinforcement learning and verification combination. The best results are in \sethlcolor{purple1}\hl{Purple}.}
 \label{tab:main_results_dpo}
  \vspace{-0.4cm}
\end{table}

\subsection{Results with Increasing Graph Size (RQ4)}
\begin{figure}[!t]
    \centering
    \includegraphics[width=0.48\textwidth]{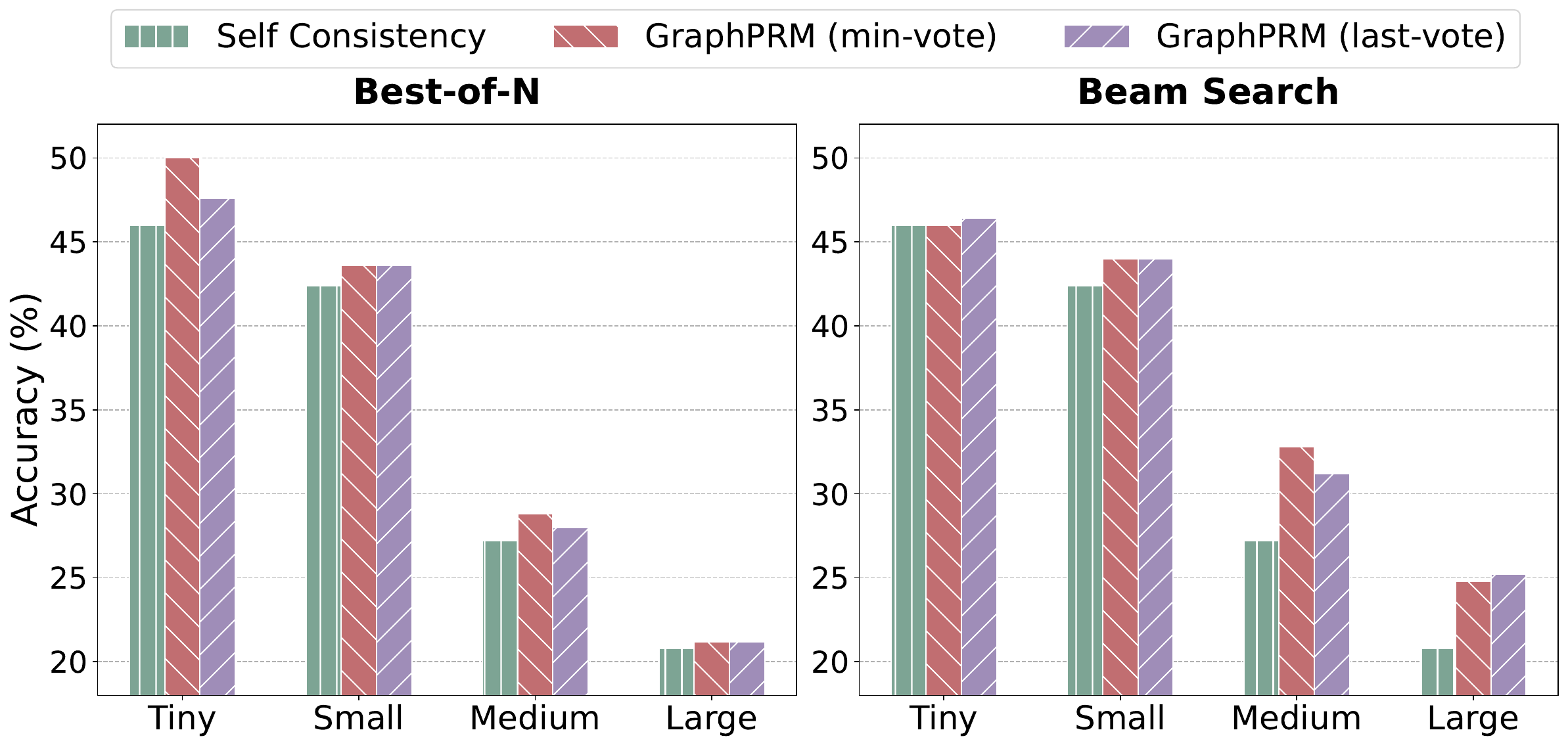}
    \caption{Performance of \OurMODEL{} on \OurDataset{} with various graph sizes. The policy model is Qwen2.5-7B-Instruct and the verification is based on 8 candidates.}
    \label{fig:graph_sizes_bar}
    \vspace{-0.2cm}
\end{figure}

To further investigate how graph sizes of test graph problems affect the performance of \OurMODEL{} on backbone LLMs, we generate four different test sets in various sizes: Tiny, Small, Medium and Large, each with 300 problems. Results on these four test sets are shown in Figure~\ref{fig:graph_sizes_bar}. It can be shown that for all inference-time searching strategies, accuracy declines with a decrease percentage around 53\% as graph size increases. Self Consistency starts with relatively high accuracy for Tiny graphs but shows a sharp decline for Medium and Large graphs. On the other hand, \OurMODEL{} with Best-of-N and Beam Search methods exhibit greater robustness, especially for Small and Medium graphs, maintaining relatively higher and more consistent accuracy compared to Self Consistency. Comparing performance among three methods on each graph size, \OurMODEL{} consistently performs beyond Self Consistency. 

Beyond this, it is evident that with the increase of graph size, \OurMODEL{} have more improvement on policy LLMs under Beam Search strategy, while have less improvement on policy LLMs under Best-of-N strategy. This decline can likely be attributed to two main factors: 1) As graph size increases, the complexity of the problem grows, the probability of the fixed candidate pool containing optimal or near-optimal solutions decreases, making it harder for Best-of-N to identify optimal solutions. 2) For larger graphs, the search space expands significantly, and the complexity of finding the best solution grows. Beam Search, supported by \OurMODEL{}, can better adapt to this complexity by incorporating voting mechanisms to systematically narrow down the search space.
This observation underscores the effectiveness and robustness of \OurMODEL{} over different sizes of graphs.

\subsection{Effect of Different Process Labeling Strategies (RQ5)}

To better understand the effectiveness of our proposed process annotation approaches discussed in Section~\ref{sec:process_annotation}, we perform an ablation study on the quality of process annotations in \OurDataset{}. As shown in Table~\ref{tab:prm_label_ablation}, \OurMODEL{} benefits significantly from both types of high-quality process annotations. When comparing the two annotation approaches, task-oriented labels yield better results under the Best-of-N searching strategy, achieving a 2.5\% improvement, whereas Monte-Carlo labels perform better under the Beam Search strategy, with a 7.2\% improvement. We identify the main rationales behind this phenomenon are: 1) Task-oriented labels are carefully designed to match the graph problem's structure and provide clear, step-by-step logical guidance. It aligns well with the static, fixed-pool nature of Best-of-N; 2) Monte-Carlo labels excel in the Beam Search strategy because their diversity and variability allow the iterative refinement process to explore a wider solution space, maximizing the potential for improvement; 3) Combining both comprehensive reasoning processes with fine-grained step-level labels from these two approaches further enhance the effectiveness of \OurMODEL{}.

\subsection{Influence of Pre-trained Model Sizes (RQ6)}

\begin{figure*}[!t]
    \centering
    \includegraphics[width=\textwidth]{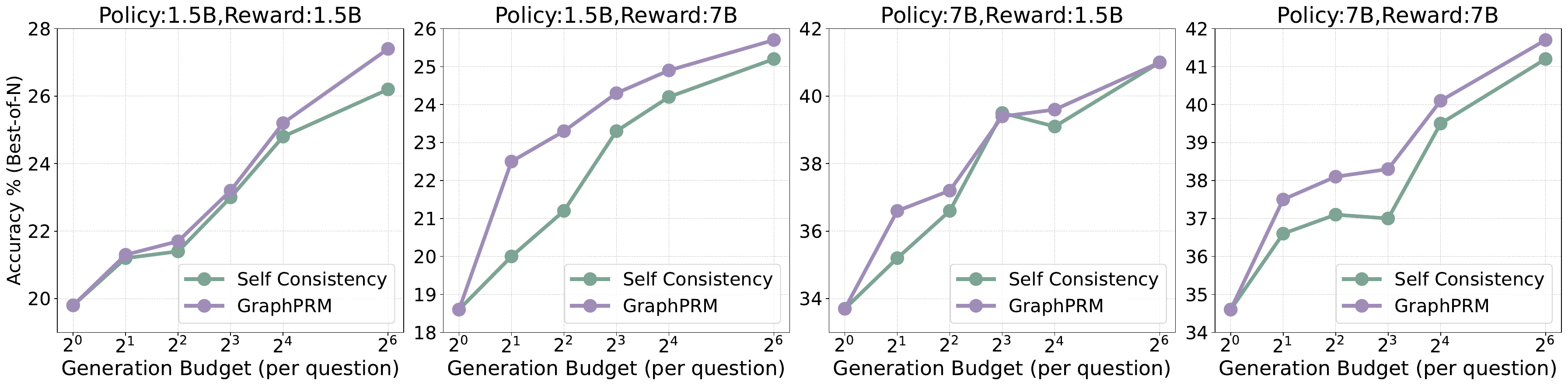}
    \caption{Best-of-N Performance of \OurMODEL{} on \OurDataset{} with policy models and reward models in different parameter sizes. The answer aggregation strategy is PRM-Last.}
    \label{fig:model_size_combine_line_plot}
\end{figure*}

To conduct an exhaustive analysis of to what extent does the combination of model parameters affect the final performance on graph reasoning tasks, we performed a range of experiments using model sizes in 1.5B and 7B. Figures~\ref{fig:model_size_combine_line_plot}(a) and \ref{fig:model_size_combine_line_plot}(d) display the results from the 1.5B and 7B policy models paired with equal-sized \OurMODEL{}s, respectively. It becomes evident that \OurMODEL{} exhibits superiority over Self Consistency across all sizes of base models.

Figure~\ref{fig:model_size_combine_line_plot}(b) and \ref{fig:model_size_combine_line_plot}(c) presents the performance of 1.5B and 7B policy models interfaced with different-sized \OurMODEL{}s. The findings illustrate that utilizing a larger \OurMODEL{} to validate the output of a smaller policy model significantly enhances performance. Conversely, when a smaller \OurMODEL{} is employed to validate the output of a larger policy model, the verification process adversely impacts the model’s performance compared to Self Consistency. These results substantiate that we should utilize a more potent \OurMODEL{} for validating or supervising the policy model.

\begin{table}[!t]\small
\renewcommand{\arraystretch}{1.1}
 \centering
 \setlength{\tabcolsep}{0.8mm}
 \resizebox{\linewidth}{!}{
 \begin{tabular}{p{0.6\linewidth}|c|c}\toprule
    \textbf{Methods} & \textbf{Best-of-N} & \textbf{Beam Search} \\
    \midrule \specialrule{0em}{1.5pt}{1.5pt}
    Qwen2.5-7B-Instruct (Self Consistency)     & 37.03 & 37.03 \\
    \midrule
    + \OurMODEL{}     & \best{38.30} & \best{41.57} \\
    + \OurMODEL{} - Task-oriented Labels     & 37.96 & 36.03 \\
    + \OurMODEL{} - Monte-carlo Labels     & 37.36 & 39.70 \\
    \bottomrule
 \end{tabular}}
 \caption{Performances (\%) of \OurMODEL{} trained on different set of step-wise process labels. The verification is based on 8 candidates. The best results are in \sethlcolor{purple1}\hl{Purple}.}
 \label{tab:prm_label_ablation}
  \vspace{-0.3cm}
\end{table}

\subsection{Case Study}

In this subsection, we present two detailed cases on both graph reasoning problem and mathematical reasoning problem, in which we compare the correct and wrong reasoning processes generated from policy LLM model, alongside with scores assigned by \OurMODEL{}. 

\begin{figure}[!htbp]
    \centering
    \includegraphics[width=0.48\textwidth]{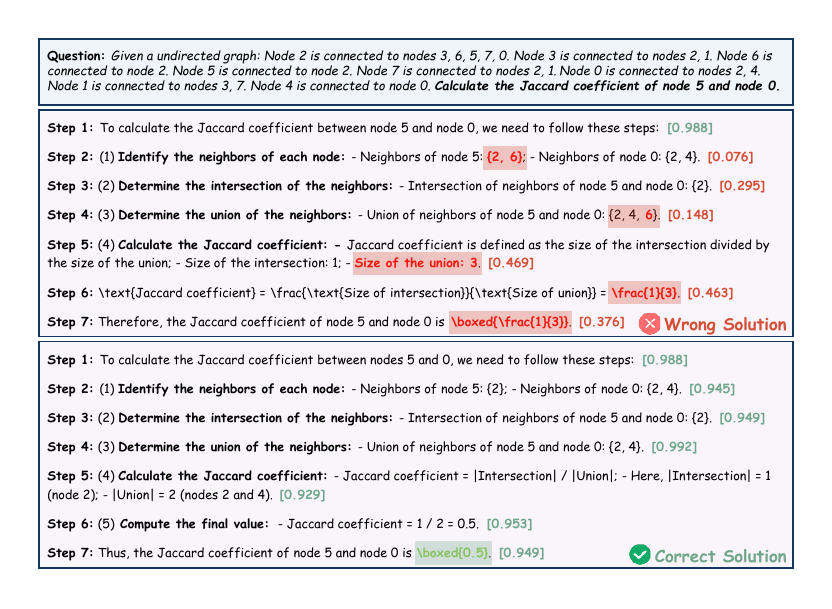}
    \caption{A case of Clustering Coefficient task in \OurDataset{}. Correct and wrong solutions are identified with \OurMODEL{} via Best-of-N strategy.}
    \label{fig:case_study_graphsilo}
    \vspace{-0.2cm}
\end{figure}

\vpara{Case on Graph Reasoning} Figure~\ref{fig:case_study_graphsilo} shows a case study of a Clustering Coefficient problem in \OurDataset{} dataset.
It can be observed that when presented with a graph reasoning problem, our \OurMODEL{} accurately selected the correct solution from a pool of candidate solutions. Specifically, in wrong solution case, LLM starts fall in wrong in step 2 when calculating the neighbors of node 5 (regarding 6 as its neighbor), thus leading to subsequent mistakes in reasoning steps. Our \OurMODEL{} displays superior discernment by precisely identifying wrong steps and assigns low scores from step 2 to step 7. In correct solution case, \OurMODEL{} precisely rank high scores to correct steps. This case study further demonstrate the effectiveness of \OurMODEL{}.

\begin{figure}[!htbp]
    \centering
    \includegraphics[width=0.48\textwidth]{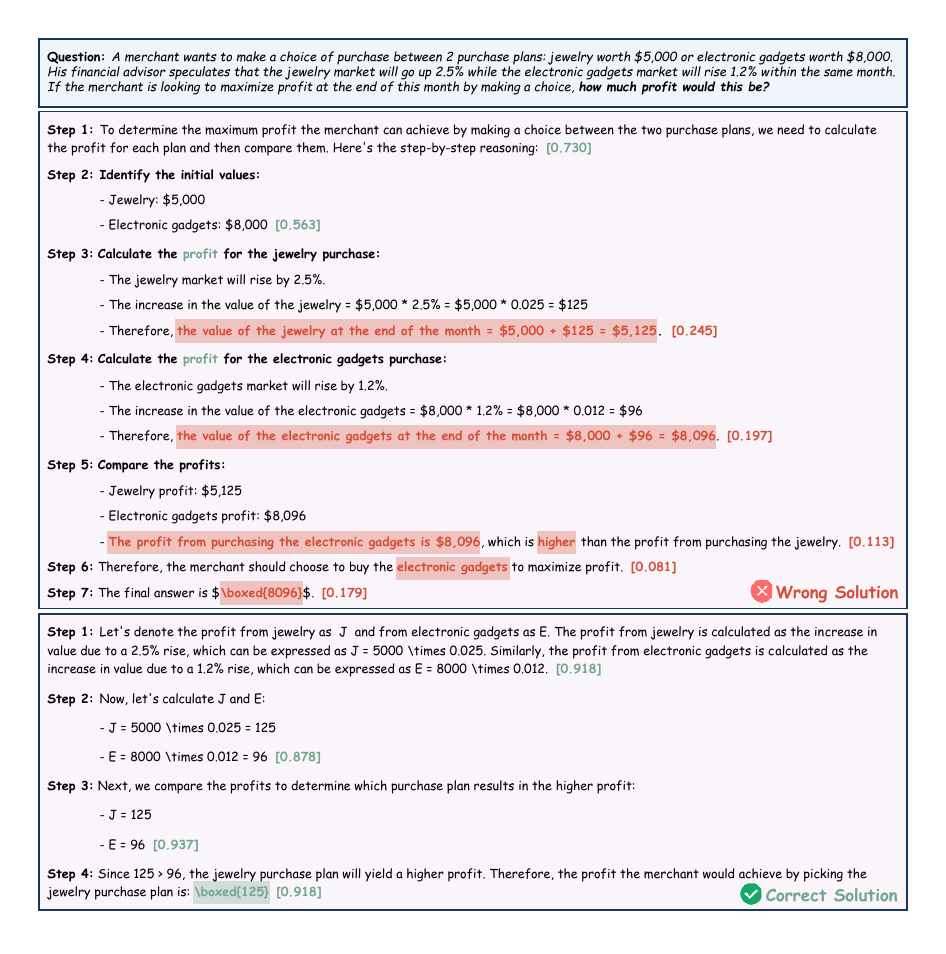}
    \caption{A case of mathematical reasoning result in GSM8K dataset. Correct and wrong solutions are identified with \OurMODEL{} via Best-of-N strategy.}
    \label{fig:case_study_GSM8K}
\end{figure}

\vpara{Case on Mathematical Reasoning} Figure~\ref{fig:case_study_GSM8K} presents a case study of a mathematical reasoning problem from the GSM8K dataset. The results in this figure highlight the effectiveness of \OurMODEL{} in verifying the correctness of the mathematical reasoning process, ultimately yielding a correct final answer through reliable step-by-step reasoning. Notably, we observe that the LLM begins to falter in step 3, where it calculates the profit for the jewelry purchase. At this step, it incorrectly interprets "the total value of the electronic gadget" as the "profit", despite performing the calculation in this step correctly. \OurMODEL{} successfully identifies this misunderstanding of the calculation target and assigns a low score to this step, along with subsequent steps that depend on the erroneous reasoning from step 3. Conversely, in the correct solution, \OurMODEL{} assigns high scores to the accurate steps. This case study further demonstrates the adaptability and transferability of \OurMODEL{} for mathematical reasoning tasks.

\section{Conclusion}

In this paper, we introduce a novel approach to enhance reasoning capabilities in LLMs by leveraging PRMs for graph reasoning tasks. We developed \OurDataset{}, the largest dataset for graph reasoning with fine-grained, step-wise labels. Building upon this dataset, we proposed \OurMODEL{}, the first PRM tailored for reasoning on GCPs, and demonstrated its effectiveness through extensive evaluations in settings such as inference-time scaling and reinforcement learning via Direct Preference Optimization (DPO). Our experiments show that \OurMODEL{} delivers significant improvements across 13 graph reasoning tasks and demonstrates strong generalization to unseen graph reasoning datasets. Notably, we observed a compelling transferability of \OurMODEL{} to other reasoning domains, including mathematical problem-solving tasks like GSM8K and MATH500. These results underscore the potential of PRMs to bridge reasoning domains and highlight their promise in advancing step-wise reasoning, laying the groundwork for applying graph-based reasoning to broader challenges and advancing LLM capabilities.

\section{Acknowledgements}
We would like to thank all the anonymous reviewers and area chairs for their insightful and valuable comments. We also thank the support of the 12th Baidu Scholarship. This research is supported by National Key Research and Development Program of China Grant No.2023YFF0725100 and Guangdong S\&T Program C019.

\bibliographystyle{ACM-Reference-Format}
\bibliography{sample-base}

\appendix

\section{More Dataset Details}

\subsection{Graph Structure Details}\label{app:graph_structure_details}
In Section~\ref{sec:graph_generation}, we elaborate on the diverse graph structure distributions employed in the construction of \OurDataset{}. Specifically, we incorporate three fundamental network topologies:
\begin{itemize}
    \item \textbf{Random networks}: These networks are constructed by establishing edges between $N$ nodes with probability $p$, resulting in a stochastic connectivity pattern.
    \item \textbf{Scale-free networks}: Characterized by a power-law degree distribution, these networks are prevalent in various real-world systems, including social interactions, biological processes, and economic trade relationships.
    \item \textbf{Small-world networks}: These networks exhibit a distinctive property where, despite relatively sparse direct connections, any two nodes can typically be connected through a small number of intermediate nodes, facilitating efficient information flow across the network.
\end{itemize}

\subsection{Dataset Statistics}
Table~\ref{tab:Benchmark Statistics} presents the dataset statistics for \OurDataset{}, providing detailed information on the PRM Training Set, PRM Test Set, SFT Training Set, and DPO Training Set for better understanding.

\begin{table}[!h]
 \centering
 \resizebox{\linewidth}{!}{
 \small
 \begin{tabular}{l|c}\toprule
    \textbf{Property} & \textbf{Number}\\
    \midrule
    \multicolumn{1}{l}{\textbf{\textit{PRM Training Set}}} \\
    \# of graph tasks & 13 \\
    \# of problems & 20,063 \\
    \# of problem-solution pairs (total) & 118,189 \\
    \# of problem-solution pairs (Task-oriented Trajectory) & 75,852 \\
    \# of problem-solution pairs (Monte-carlo Method) & 42,337 \\
    \# of positive step labels & 220,894 \\
    \# of negative step labels & 173,271 \\
    \# of maximum steps in a solution & 20 \\
    \midrule
    \multicolumn{1}{l}{\textbf{\textit{PRM Test Set}}} \\
    \# of problems (total) & 3,900 \\
    \# of problems per task & 300 \\
    \# of graph size types & 4\\
    \midrule
    \multicolumn{1}{l}{\textbf{\textit{SFT Training Set}}} \\
    \# of problem-solution pairs (total) & 7,964 \\
    \midrule
    \multicolumn{1}{l}{\textbf{\textit{DPO Training Set}}} \\
    \# of positive-negative preference pairs (origin Qwen) & 6,137 \\
    \# of positive-negative preference pairs (SFT model) & 5,031 \\
    \bottomrule
 \end{tabular}}
 \caption{Data Statistics of \OurDataset{}}
 \label{tab:Benchmark Statistics}
  \vspace{-0.3cm}
\end{table}

\subsection{Quality Control for \OurDataset{}}

Ideally step-wise process labels should be clear enough to support the validity of corresponding solution sub-steps. After obtaining graph reasoning processes with step-level labels via approaches proposed in Section~\ref{sec:process_annotation}, we conduct quality control pipeline to select high-quality data for further analysis. Specifically, we include the following three steps:

\vpara{Completeness Filtering} All reasoning processes in \OurDataset{} should be meaningful and related to corresponding graph problem. For data generated from "Task-oriented trajectories" approach, their completeness and correctness are inherently assured due to the advantages of graph algorithms. For data generated from "Automated Monte-Carlo Method" approach, we filter out those: 1) reasoning processes without a final answer; 2) reasoning processes containing repeated meaningless sentences.

\vpara{Label Filtering} Each step in a solution is assigned a binary label, `+' for positive and `-' for negative. For problem-solution pairs generated using the "Task-oriented Trajectories" method, we retain only those pairs where negative steps occur exclusively after positive steps. For pairs generated via the "Automated Monte-Carlo Method," all steps following an incorrect step (label `-') are removed, as the validity of correct step becomes irrelevant after an error. This removal is designed to avoid potential confusion during model training~\cite{ProcessBench}.

\vpara{Human Validation} We recruited five volunteers to evaluate the entailment between step-wise label and corresponding reasoning sub-step. Each volunteer assessed a sample of 500 randomly selected examples to ensure the quality and reliability of our dataset. They were tasked with "Determining whether the step-wise label supports the corresponding reasoning step regrading the graph problem." The human evaluation results are listed in Table~\ref{tab:quality_control}. The high agreement observed further supports our dataset's quality and validity of step-wise process labels.

\begin{table}[!htbp]
 \centering
 \resizebox{\linewidth}{!}{
 \small
 \begin{tabular}{l|c|c|c|c|c}\toprule
     & \textbf{Anno-1} & \textbf{Anno-2} & \textbf{Anno-3} & \textbf{Anno-4} & \textbf{Anno-5}\\
    \midrule
    Agreement Rate & 95.0 & 94.2 & 95.4 & 96.0 & 94.4\\
    \bottomrule
 \end{tabular}}
 \caption{Human evaluation results on \OurDataset{}}
 \label{tab:quality_control}
  \vspace{-0.3cm}
\end{table}

\subsection{Detailed Task Definition}\label{sec:detailed_task_definition}

In this section, we present detailed definitions for each task included in the \OurDataset{} dataset. For tasks in Boolean output format, we curate a balanced dataset by filtering the graphs to ensure an equal distribution of positive and negative labels.
\begin{itemize}[leftmargin=*]
    \item \textbf{Task 1: Degree.} Consider a graph $\mathcal{G}=(\mathcal{V},\mathcal{E})$, which may be either directed or undirected. For a specified node $v \in \mathcal{V}$, compute its degree: in an undirected graph, the degree is the number of incident edges; in a directed graph, determine both the in-degree (number of incoming edges) and the out-degree (number of outgoing edges).

    \item \textbf{Task 2: Clustering Coefficient.} Consider a graph $\mathcal{G}=(\mathcal{V},\mathcal{E})$, which may be directed or undirected. For a specified node $u\in\mathcal{V}$, the task is to compute its clustering coefficient as follows:
    
    \begin{itemize}
        \item For a \textbf{directed} graph, let $T$ be the number of edges among the out-neighbors of $u$ and $D$ be the out-degree of $u$. The clustering coefficient is then given by
        $C = \frac{T}{D(D - 1)}$.
        \item For an \textbf{undirected} graph, using the same definitions for $T$ and $D$, the clustering coefficient is defined as
        $C = \frac{2T}{D(D - 1)}$.  
    \end{itemize}

The objective is to accurately calculate and report the clustering coefficient value for node $u$ based on the type of graph provided.

    \item \textbf{Task 3: Neighbor.} Consider a graph $\mathcal{G}=(\mathcal{V},\mathcal{E})$ that may be either directed or undirected. For a given node $u \in \mathcal{V}$, identify its neighbors—that is, all nodes directly connected to $u$. In an undirected graph, these are all nodes $v$ with ${u,v}\in\mathcal{E}$; in a directed graph, they include both nodes $v$ with $(u,v)\in\mathcal{E}$ (out-neighbors) and nodes $v$ with $(v,u)\in\mathcal{E}$ (in-neighbors). 

    \item \textbf{Task 4: Page Rank.} Compute the PageRank values for each node in graph $\mathcal{G}=(\mathcal{V},\mathcal{E})$ (directed or undirected) by initializing each node's PageRank to $1/|\mathcal{V}|$, then performing 3 iterations of the algorithm with a damping factor of 0.85. Finally, identify the node with the largest PageRank value.

    \item \textbf{Task 5: Predecessor.} Consider a graph $\mathcal{G}=(\mathcal{V},\mathcal{E})$ that may be either directed or undirected. For a given node $u$, return its predecessor nodes as follows: in a directed graph, a predecessor of $u$ is any node $m$ such that there is an edge from $m$ to $u$; in an undirected graph, treat all neighbors of $u$ as its predecessors.

    \item \textbf{Task 6: Jaccard.} Consider a graph $\mathcal{G}=(\mathcal{V},\mathcal{E})$ that may be either directed or undirected. For two given nodes $u$ and $v$, compute their Jaccard similarity, defined as the size of the intersection of their neighbor sets divided by the size of the union of their neighbor sets. In an undirected graph, the neighbor set of a node consists of all nodes directly connected to it, while in a directed graph, it includes both in- and out-neighbors. Return the computed Jaccard similarity value for the pair $(u,v)$.
   
    \item \textbf{Task 7: Common Neighbor.} Consider a graph $\mathcal{G}=(\mathcal{V},\mathcal{E})$ that may be either directed or undirected. For two given nodes $u$ and $v$, identify the common neighbors—that is, the nodes that are connected to both $u$ and $v$. In a directed graph, treat the union of in-neighbors and out-neighbors as the neighbor set for each node.
    
    \item \textbf{Task 8: Connectivity.} Given graph $\mathcal{G}=\{\mathcal{V},\mathcal{E}\}$ that may be either directed or undirected. the task is to assess if two randomly chosen nodes $u$ and $v$ are connected through a sequence of edges. 

    \item \textbf{Task 9: Maximum Flow.} Consider a directed, weighted graph $\mathcal{G}=\{\mathcal{V},\mathcal{E}, c\}$, 
    where $c: \mathcal{E} \to \mathbb{R}^+$ is a function assigning a positive capacity to each edge, representing the maximum flow that the edge can support. Given a source node $v_s$ and a sink node $v_t$ in $\mathcal{G}$, the task is to devise a plan to maximize the flow from the source $s$ to the sink $t$. 

    \item \textbf{Task 10: Breadth First Search.} Consider a graph $\mathcal{G}=(\mathcal{V},\mathcal{E})$ that may be either directed or undirected. For a given starting node $u$, perform a BFS traversal and return the sequence of nodes in the order they are visited. In a directed graph, use out-neighbors as adjacent nodes; in an undirected graph, use all directly connected neighbors.
    
    \item \textbf{Task 11: Cycle.} In a graph $\mathcal{G}=\{\mathcal{V},\mathcal{E}\}$, the task is to detect the existence of a cycle. A cycle can be defined as a sequence of vertices $v_1, v_2, \dots, v_k$ with $k\geq 3$, that forms a closed loop, meaning $v_1=v_k$. Additionally, for all $1\leq i < k$, each vertex $v_i$ must be distinct from the others, and there must be an edge connecting $v_i$ to $v_{i+1}$. 
    
    \item \textbf{Task 12: Diameter.} Consider a graph $\mathcal{G}=(\mathcal{V},\mathcal{E})$, which may be either directed or undirected. The task is to compute the diameter of $\mathcal{G}$, defined as the maximum shortest path distance between any two nodes. Return the computed diameter.
    
    \item \textbf{Task 13: Minimum Spanning Tree.} Consider an undirected, weighted graph $\mathcal{G}=(\mathcal{V},\mathcal{E},w)$, where $w$ assigns a weight to each edge. The task is to compute the Minimum Spanning Tree (MST) of $\mathcal{G}$—a subset of edges that connects all nodes with the minimum total edge weight. Return the MST or its total weight.

\end{itemize}

\section{Additional Experimental Results}\label{app:additional_experimental_results}
In this section, to further analysis our proposed \OurMODEL{}, we evaluate \OurMODEL{} by addressing the following extra research questions:

\begin{itemize}
\item \textbf{RQ7:} How does \OurMODEL{} perform on other GCP-specific backbones?
\item \textbf{RQ8:} How does graph density affect the performance of \OurMODEL{} on LLMs during inference time?
\end{itemize}

\subsection{Performance of \OurMODEL{} on GraphWiz model (RQ7)}
\begin{figure}[!t]
    \centering
    \includegraphics[width=0.46\textwidth]{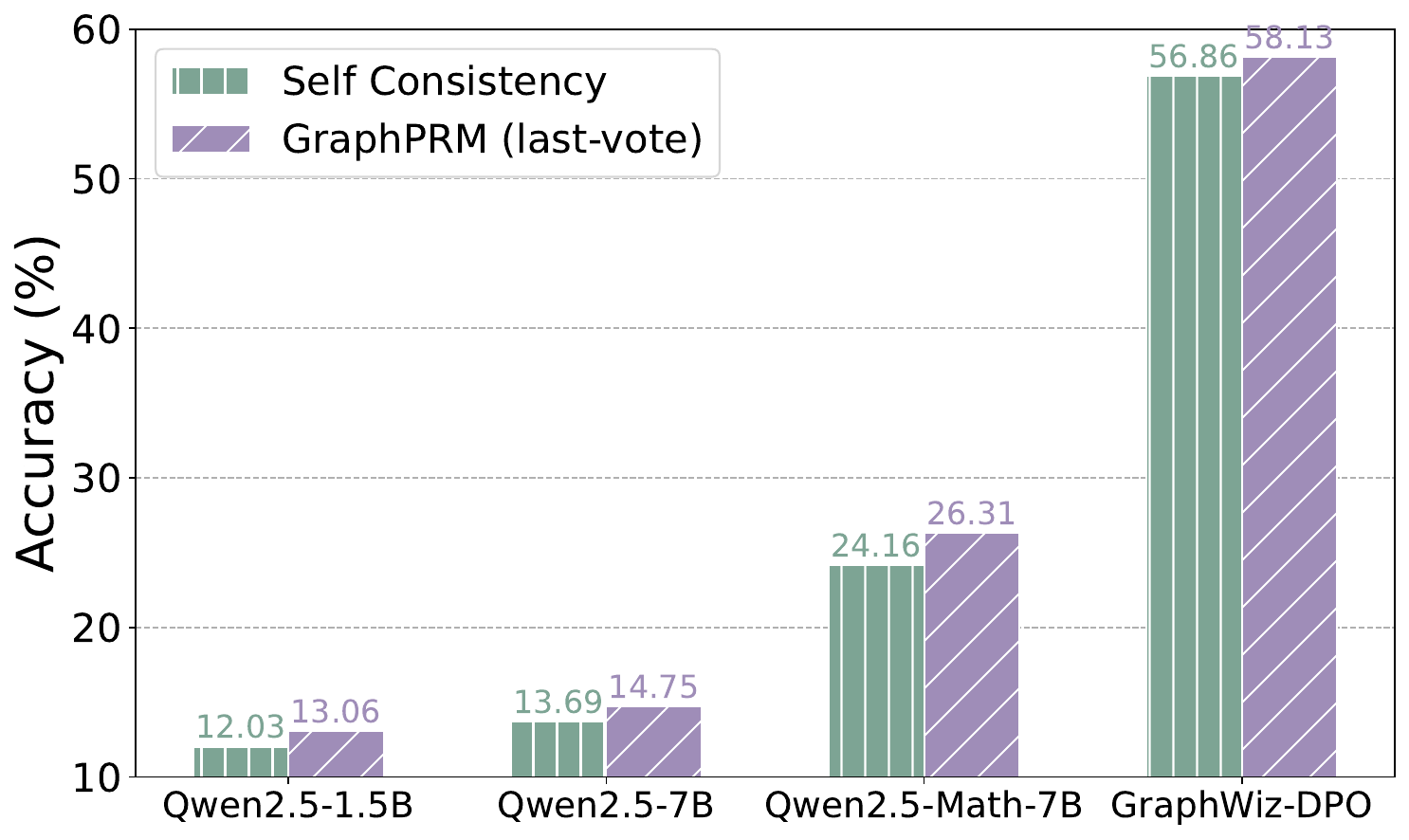}
    \caption{Best-of-N Performance of \OurMODEL{} on GraphWiz with different backbones. We adopt GraphWiz-DPO trained on LLaMA2-7B. Verification is based on 8 candidates.}
    \label{fig:graphwiz_backbones_bar}
\end{figure}

In addition to our proposed \OurDataset{}, we conduct experiments on GraphWiz dataset with Qwen2.5-7B-Instruct, Qwen2.5-Math-7B-Instruct and LLaMA-3.1-8B-Instruct in Section~\ref{sec:performance_of_GraphPRM}. Besides above, we further evaluate the transferbility of \OurMODEL{} on other GCP-specific model like GraphWiz. As shown in Figure~\ref{fig:graphwiz_backbones_bar}, \OurMODEL{} consistently improve LLMs' performance on GraphWiz dataset with more than 5\% improvement.

\subsection{Impact of Graph Density (RQ8)}
\begin{figure}[!t]
    \centering
    \includegraphics[width=0.46\textwidth]{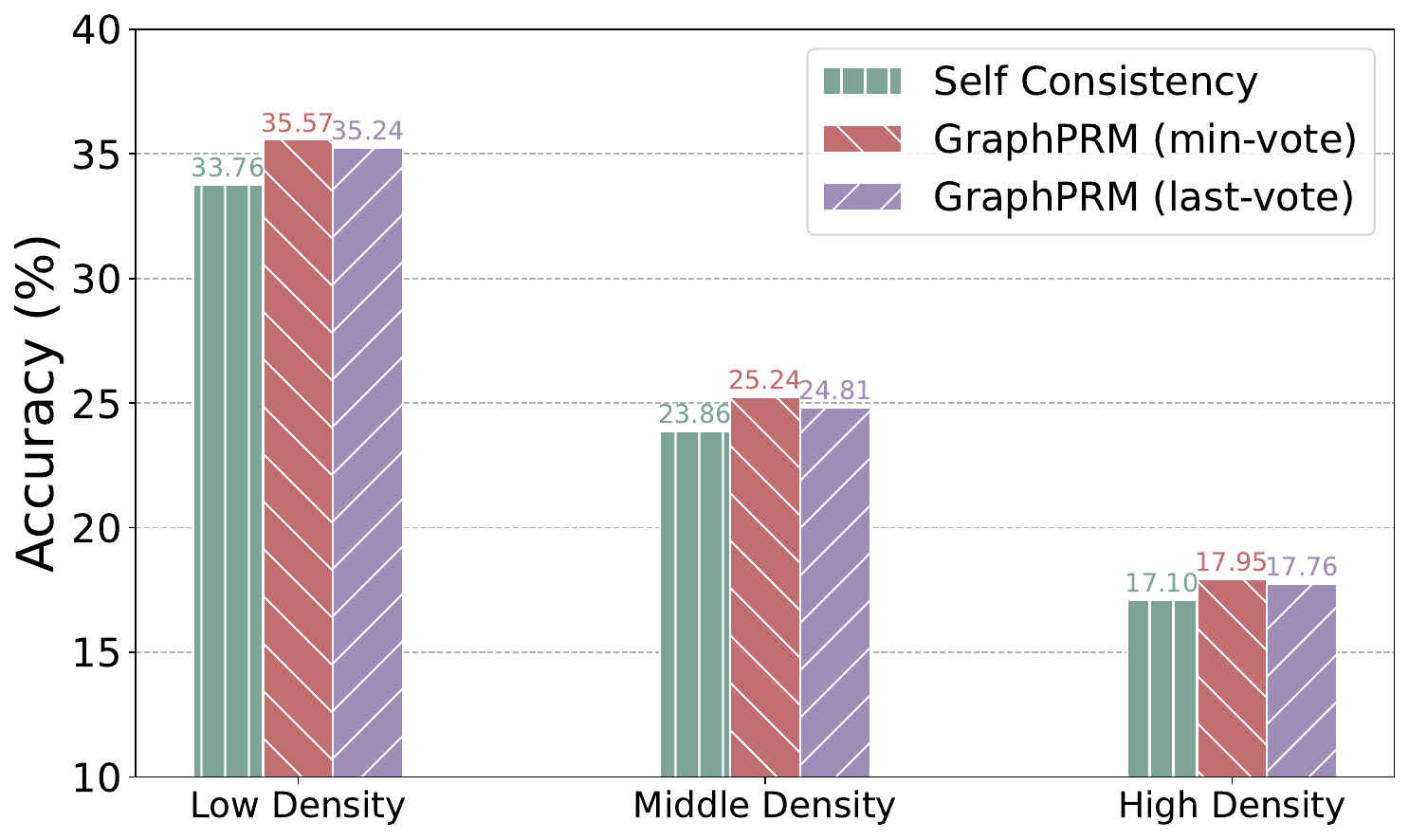}
    \caption{Best-of-N Performance of \OurMODEL{} on \OurDataset{} with different graph densities. The policy model is Qwen2.5-7B-Instruct and the verification is based on 8 candidates.}
    \label{fig:graph_density_bar}
\end{figure}

Graph density can be a crucial factor influencing the performance of \OurMODEL{} on graph reasoning tasks. We regard \textbf{Edge Density} $\left(x = \frac{2|E|}{n(n-1)}\right)$ as the reflection of graph density, and we construct new test set with three-level density: Low ($0 < x \leq 0.33$), Middle ($0.33 < x \leq 0.67$) and High ($0.67 < x \leq 1$). Results with different graph densities are shown in figure~\ref{fig:graph_density_bar}. It can be observed that accuracy decreases as graph density increases, but the model with \OurMODEL{} consistently outperforms that with Self-Consistency, demonstrating its robustness across graphs with different densities. Besides, as graph density increases, the performance gained via \OurMODEL{} (Best-of-8) decreases. We conclude this is the reason that: Graphs with higher density contain more invalid edges or noisy edges, bringing more challenges for GraphPRM to identify essential steps. This further highlights the importance of employing \OurMODEL{} to supervise step-wise graph reasoning.

\section{More Cases}

\subsection{Task Prompts Used in Experiments}
In this section, we provide a detailed list of all prompts for all experiments, offering a clear reference for understanding our experimental approach. Prompts for evaluating graph reasoning tasks in \OurDataset{} on both inference-time computation and reinforcement learning settings are listed in Table~\ref{tab:prompts_graph_tasks}.

\begin{table*}[t!]
\renewcommand{\arraystretch}{1.2}
\resizebox{\linewidth}{!}{
    \centering
\begin{tabular}{l|p{0.8\linewidth}}
\toprule
\textbf{Task} & \textbf{Prompt} \\
\midrule
\textbf{Degree} & Given a directed/undirected graph:\textcolor[RGB]{158,160,161}{[Graph edgelist in natural language]} What is the degree of node 2? Please reason step by step, and put your final answer within \textbackslash boxed\{\}.\\ 
\midrule
\textbf{Clustering Coefficient} & Given a directed/undirected graph:\textcolor[RGB]{158,160,161}{[Graph edgelist in natural language]} What is the clustering coefficient of node 14? For a directed graph, we consider a node's successors as its neighbors. Please reason step by step, and put your final answer within \textbackslash boxed\{\}. \\ 
\midrule
\textbf{Neighbor} & Given a directed/undirected graph:\textcolor[RGB]{158,160,161}{[Graph edgelist in natural language]} Which are the neighbor nodes of node 2? Please reason step by step, and put your final answer within \textbackslash boxed\{\}. The answer should be in the form of an array that starts with'[' and ends with ']', separated by comma ','. \\ 
\midrule
\textbf{Page Rank} & Given a directed/undirected graph:\textcolor[RGB]{158,160,161}{[Graph edgelist in natural language]} Which node has the largest PageRank value? The dampling factor is 0.85. The number of iterations is 3. The initial PageRank values for all nodes are initialized equally as 1/N, where N is the number of nodes. Please reason step by step, and put your final answer within \textbackslash boxed\{\}.\\ 
\midrule
\textbf{Predecessor} & Given a directed/undirected graph:\textcolor[RGB]{158,160,161}{[Graph edgelist in natural language]} Which are the predecessor nodes of node 8? A predecessor of n is a node m such that there exists a directed edge from m to n. Please reason step by step, and put your final answer within \textbackslash boxed\{\}. The answer should be in the form of an array that starts with'[' and ends with ']', separated by comma ','.\\ 
\midrule
\textbf{Jaccard} & Given a directed/undirected graph:\textcolor[RGB]{158,160,161}{[Graph edgelist in natural language]} Calculate the Jaccard coefficient of node 0 and node 3. For a directed graph, we consider a node's successors as its neighbors. Please reason step by step, and put your final answer within \textbackslash boxed\{\}.\\ 
\midrule
\textbf{Common Neighbor} & Given a directed/undirected graph:\textcolor[RGB]{158,160,161}{[Graph edgelist in natural language]} Calculate the number of common neighbors of node 2 and node 9. In the context of a directed graph, we consider a node's successors as its neighbors. Please reason step by step, and put your final answer within \textbackslash boxed\{\}.\\ 
\midrule
\textbf{Connectivity} & Given a directed/undirected graph:\textcolor[RGB]{158,160,161}{[Graph edgelist in natural language]} Is there a path between node 0 and node 2? Please reason step by step, and answer with \textbackslash boxed\{True\} or \textbackslash boxed\{False\}. \\ 
\midrule
\textbf{Maximum Flow} & Given a directed/undirected graph:\textcolor[RGB]{158,160,161}{[Graph edgelist in natural language]} Calculate the maximum flow between node 2 and node 11 in this graph. Given a directed graph with capacities assigned to its edges, the maximum flow from a source node to a sink node is the maximum amount of flow that can be sent from the source to the sink, respecting the capacity constraints on each edge. The goal is to find the optimal way to route flow through the network to maximize the flow from source to sink. Please reason step by step, and put your final answer within \textbackslash boxed\{\}.\\ 
\midrule
\textbf{Breadth First Search} & Given a directed/undirected graph:\textcolor[RGB]{158,160,161}{[Graph edgelist in natural language]} Start from node 7, output a sequence of traversal in breadth-first search (BFS) order. Please reason step by step, and put your final answer within \textbackslash boxed\{\}. The answer should be in the form of an array that starts with'[' and ends with ']', separated by comma ','\\ 
\midrule
\textbf{Cycle} & Given a directed/undirected graph:\textcolor[RGB]{158,160,161}{[Graph edgelist in natural language]} Does the graph have a cycle? For a directed graph, a cycle is a closed path that traverses through a sequence of nodes and directed edges, eventually returning to the starting node. Please reason step by step, and answer with \textbackslash boxed\{True\} or \textbackslash boxed\{False\}.\\ 
\midrule
\textbf{Diameter} & Given an undirected graph:\textcolor[RGB]{158,160,161}{[Graph edgelist in natural language]} Calculate the diameter of the graph. The diameter is the maximum distance over all pairs of nodes in the graph. Please reason step by step, and put your final answer within \textbackslash boxed\{\}.\\ 
\midrule
\textbf{Minimum Spanning Tree} & Given an undirected graph:\textcolor[RGB]{158,160,161}{[Graph edgelist in natural language]} Output the total weight of the minimum spanning tree (MST) for this graph. Please reason step by step, and put your final answer within \textbackslash boxed\{\}.\\ 
\bottomrule
\end{tabular}}
    \caption{Prompts for evaluating graph reasoning tasks in \OurDataset{} on both inference-time computation and reinforcement learning settings. For readability, we omit the input graph edge list in the format of natural language.}
    \label{tab:prompts_graph_tasks}
\end{table*}

\subsection{Cases with \OurMODEL{} Scores}
\begin{figure*}[!htbp]
    \centering
    \includegraphics[width=0.88\textwidth]{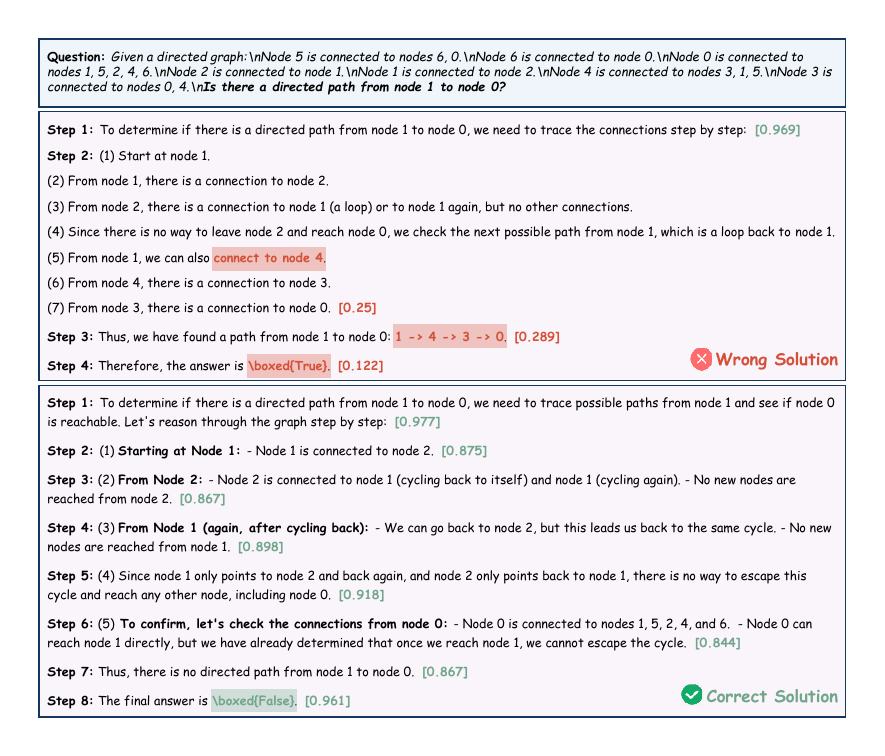}
    \caption{A case of Connectivity task in \OurDataset{}. Correct and wrong solutions are identified with \OurMODEL{} via Best-of-N strategy.}
    \label{fig:case_study_connectivity}
\end{figure*}

\begin{figure*}[!htbp]
    \centering
    \includegraphics[width=0.88\textwidth]{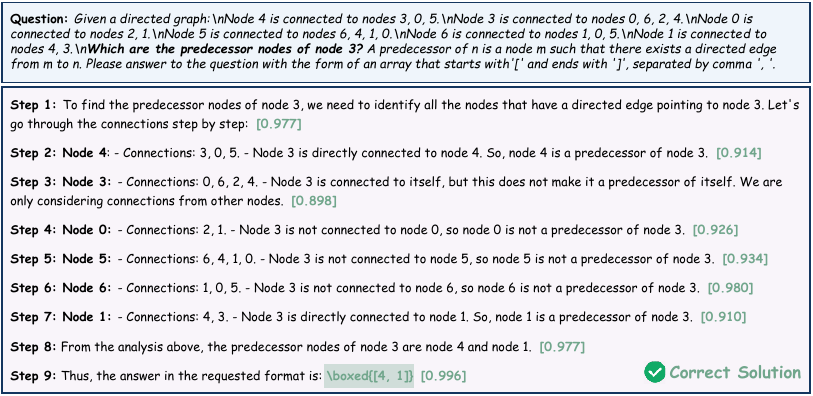}
    \caption{A case of Predecessor task in \OurDataset{}. The best solution is identified with \OurMODEL{} via Beam Search strategy.}
    \label{fig:case_study_predecessor}
\end{figure*}

\begin{figure*}[!htbp]
    \centering
    \includegraphics[width=0.88\textwidth]{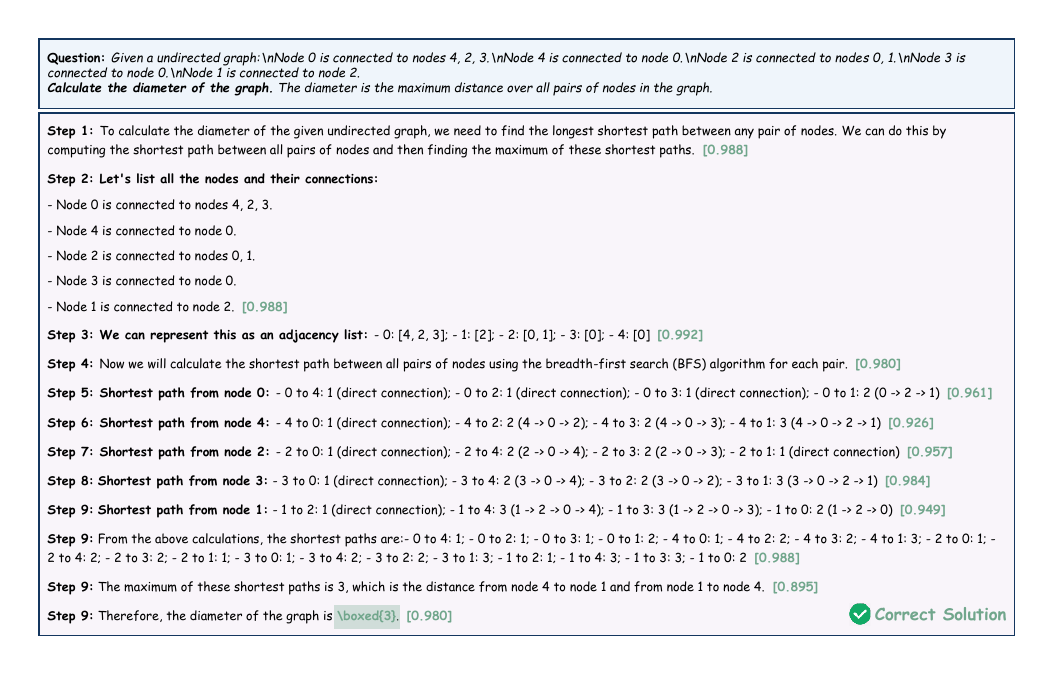}
    \caption{A case of Diameter task in \OurDataset{}. The best solution is identified with \OurMODEL{} via Beam Search strategy.}
    \label{fig:case_study_diameter}
\end{figure*}

\begin{figure*}[!htbp]
    \centering
    \includegraphics[width=0.88\textwidth]{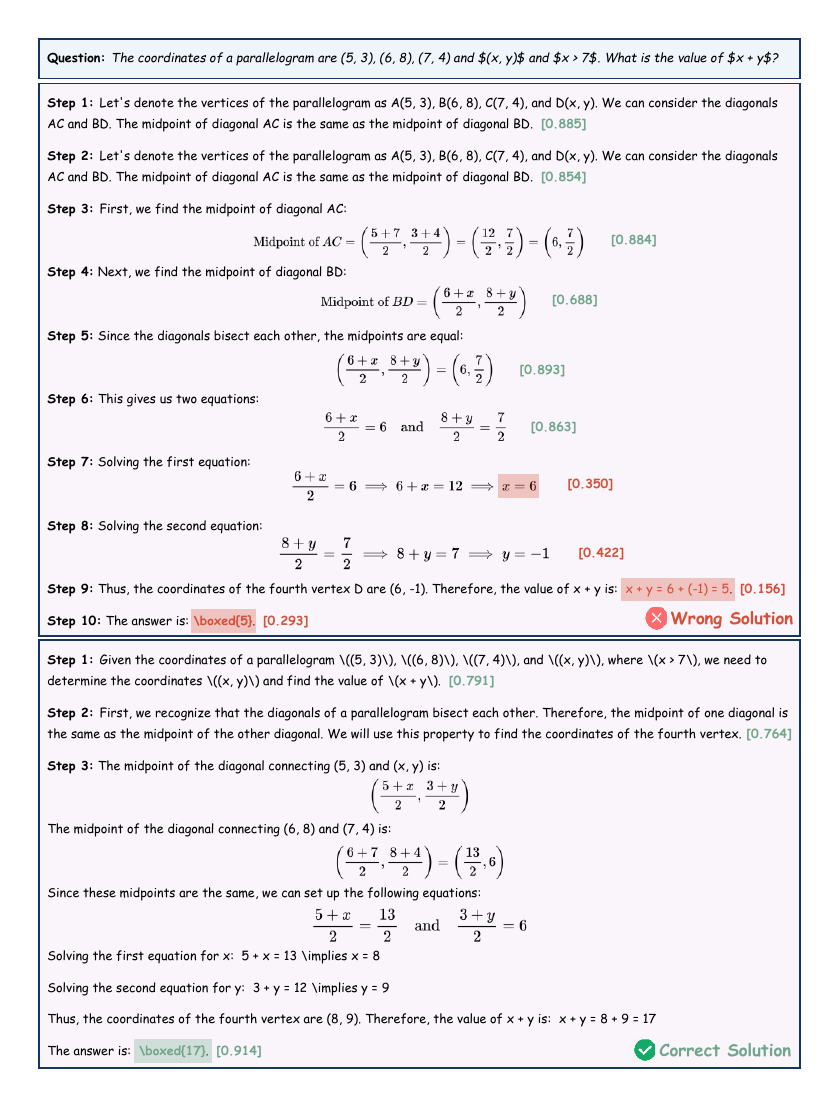}
    \caption{A case of mathematical reasoning result in MATH dataset. Correct and wrong solutions are identified with \OurMODEL{} via Best-of-N strategy.}
    \label{fig:case_study_MATH}
\end{figure*}

In this section, we provide a detailed list of reasoning cases scored by \OurMODEL{}, offering a clear reference for understanding the effectiveness of \OurMODEL{}:
\begin{itemize}
    \item Cases of reasoning results on \textit{Connectivity}, \textit{Predecessor} and \textit{Diameter} tasks in \OurDataset{} are listed in Figure~\ref{fig:case_study_connectivity}, Figure~\ref{fig:case_study_predecessor} and Figure~\ref{fig:case_study_diameter}, respectively.
    \item Cases of mathematical reasoning results on MATH dataset is listed in Figure~\ref{fig:case_study_MATH}.
\end{itemize}

\end{document}